\pdfoutput=1
\PassOptionsToPackage{bookmarks=false,breaklinks=true,letterpaper=true,colorlinks,linkcolor=black,citecolor=blue,urlcolor=black}{hyperref}
\documentclass[conference]{IEEEtran}
\IEEEoverridecommandlockouts

\usepackage[nocompress]{cite}
\usepackage{amsmath,amssymb,amsfonts}
\usepackage{graphicx}
\usepackage{textcomp}
\usepackage{xcolor}
\usepackage{notoccite}
\usepackage[english]{babel}
\usepackage[utf8x]{inputenc}
\usepackage{amsmath,amsthm}
\usepackage{multirow, array}
\usepackage{algorithm}
\usepackage[noend]{algpseudocode}
\algrenewcommand\algorithmicindent{0.7em}%

\usepackage{fancyhdr}
\usepackage[normalem]{ulem}
\usepackage[hyphens]{url}
\usepackage[final]{microtype}
\usepackage{flushend}

\usepackage{comment}
\newcommand{\para}[1]{\vspace{.05in} \noindent \textbf{#1}}

\usepackage{listings}

\ifCLASSOPTIONcompsoc
  \usepackage[caption=false,font=normalsize,labelfont=sf,textfont=sf]{subfig}
\else
  \usepackage[caption=false,font=footnotesize]{subfig}
\fi

\usepackage{etoolbox}

\makeatletter
\newcommand*{\algrule}[1][\algorithmicindent]{%
  \makebox[#1][l]{%
    \hspace*{.2em}
    \vrule height .75\baselineskip depth .25\baselineskip
  }
}

\newcount\ALG@printindent@tempcnta
\def\ALG@printindent{%
    \ifnum \theALG@nested>0
    \ifx\ALG@text\ALG@x@notext
    \else
    \unskip
    \ALG@printindent@tempcnta=1
    \loop
    \algrule[\csname ALG@ind@\the\ALG@printindent@tempcnta\endcsname]%
    \advance \ALG@printindent@tempcnta 1
    \ifnum \ALG@printindent@tempcnta<\numexpr\theALG@nested+1\relax
    \repeat
    \fi
    \fi
}
\patchcmd{\ALG@doentity}{\noindent\hskip\ALG@tlm}{\ALG@printindent}{}{\errmessage{failed to patch}}
\patchcmd{\ALG@doentity}{\item[]\nointerlineskip}{}{}{} 
\makeatother

\def\BibTeX{{\rm B\kern-.05em{\sc i\kern-.025em b}\kern-.08em
    T\kern-.1667em\lower.7ex\hbox{E}\kern-.125emX}}

\begin{document}

\title{Morph: Flexible Acceleration for 3D CNN-based Video Understanding
\thanks{This work was partially supported by NSF award CCF-1725734 and a DARPA SDH contract.}
}


\author{Kartik Hegde$^{\dagger}$, \thanks{$^{\dagger}$Authors contributed equally to this work.}
        Rohit Agrawal$^{\dagger}$,
        Yulun Yao,
        Christopher W. Fletcher\\
        University of Illinois at Urbana-Champaign\\
        \{kvhegde2, rohita2, yuluny2, cwfletch\}@illinois.edu}

\newcommand{\code}[1]{\texttt{#1}}

\newcommand{\chris}[1]{}
\newcommand{\kartik}[1]{}
\newcommand{\rohit}[1]{}
\newcommand{\jose}[1]{}

\maketitle

\thispagestyle{fancy}
\chead{Appears in the proceedings of the 51st Annual IEEE/ACM International Symposium on Microarchitecture~(MICRO), 2018}

\begin{abstract}

The past several years have seen both an explosion in the use of Convolutional Neural Networks (CNNs) and the design of accelerators to make CNN inference practical.
In the architecture community, the lion share of effort has targeted CNN inference for image recognition.
The closely related problem of \emph{video recognition} has received far less attention as an accelerator target.
This is surprising, as video recognition is more computationally intensive than image recognition, and video traffic is predicted to be the majority of internet traffic in the coming years.

This paper fills the gap between algorithmic and hardware advances for video recognition by providing a design space exploration and flexible architecture for accelerating 3D Convolutional Neural Networks (3D CNNs)---the core kernel in modern video understanding.
When compared to (2D) CNNs used for image recognition, efficiently accelerating 3D CNNs poses a significant engineering challenge due to their large (and variable over time) memory footprint and higher dimensionality.

To address these challenges, we design a novel accelerator, called \emph{Morph}, that can adaptively support different spatial and temporal tiling strategies depending on the needs of each layer of each target 3D CNN. 
We codesign a software infrastructure alongside the Morph hardware to find good-fit parameters to control the hardware.
Evaluated on state-of-the-art 3D CNNs, Morph achieves up to $3.4\times$ ($2.5\times$ average) reduction in energy consumption and improves performance/watt by up to $5.1\times$~($4\times$ average) compared to a baseline 3D CNN accelerator, with an area overhead of 5\%.
Morph further achieves a $15.9\times$ average energy reduction on 3D CNNs when compared to Eyeriss.
\end{abstract}


\section{Introduction}
\label{sec:intro}

The rise of Convolutional Neural Networks (CNNs)~\cite{alexnet, inception,resnet,vgg} has marked tremendous progress in image recognition, advancing the state-of-the-art in tasks ranging from handwritten digit~\cite{mnist} to complex object recognition~\cite{cifar10, imagenet}.
At their core, CNNs are compute intensive, parallel dot product operations.
Combined with their importance, this computation style has made CNNs a natural target for hardware ASIC acceleration, and a rich line of work has made large strides in this direction~\cite{eyeriss,cnvlutin,dadiannao,shidiannao,scnn,ucnn}.

Given the recent progress in image recognition, a natural question is whether similar strides have been made for the related problem of \emph{video recognition}.
Like image recognition, video understanding has received significant attention in the computer vision community at the algorithm level~\cite{twostream, karpathy, c3d, i3d, conv3d}, with numerous datasets being developed for different domains~\cite{ucf101, activitynet, moments, youtube8m}. 
Current state-of-the-art results are achieved using 3-dimensional (3D) CNNs, which generalize (2D) CNNs used for image recognition to account for the time dimension, thereby allowing the model to capture spatio-temporal features.
3D CNNs use a similar style of computation (i.e., parallel dot products with sliding window data access patterns) as 2D CNNs and are likewise extremely compute intensive.

Given the above, it is perhaps surprising that 3D CNNs have not yet received attention as a target for acceleration in ASICs.
Video processing is an important workload, and video traffic is predicted to account for 78\% of all internet traffic by 2021~\cite{cisco}.
Additionally, a large number of real-life applications based on video understanding, e.g., surveillance drones, self-driving cars etc., mandate real-time video understanding, further showing the need to provide hardware acceleration for 3D CNNs.

To bridge this gap, this paper studies hardware acceleration for 3D CNN inference using ASICs, in performance- and energy- constrained environments.


\begin{figure*}[!t]
    \centering
    \subfloat[Memory footprint comparison for different layers of representative 2D and 3D CNNs. 
    Assumes a $224 \times 224$ input frame with $3$ channels and $16$ frames, convolved with a $3 \times 3$ filter with $3$ channels and $3$ temporal depth.
    We compare C3D and Alexnet as they have a similar structure, but remark that more recent 2D CNNs such as ResNet have comparable input/weight footprints. 
    \label{fig:size_2dvs3d}]{
    \includegraphics[width=0.45\textwidth]{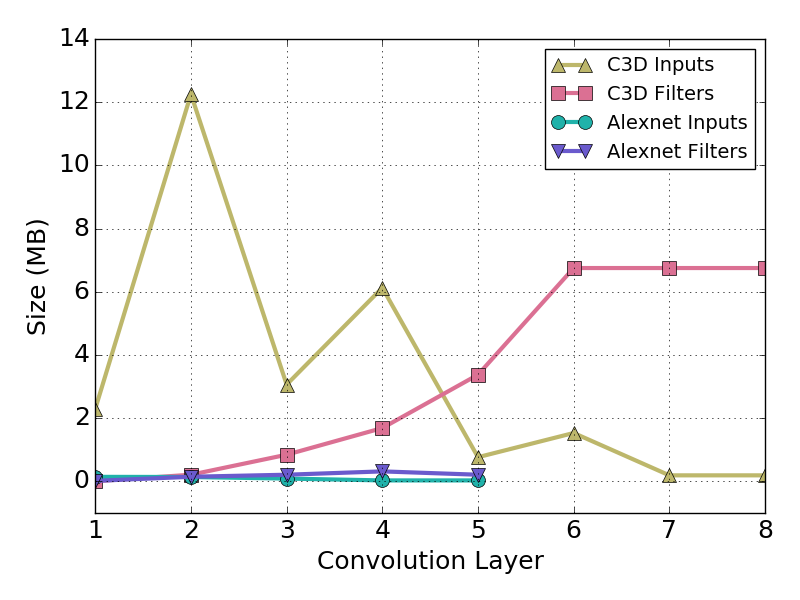}
    } ~
    \subfloat[Average data reuse in 2D and 3D CNNs.  Input activations and weights are 1~Byte each.
    \label{fig:compute_2dvs3d}]{
    \includegraphics[width=0.45\textwidth]{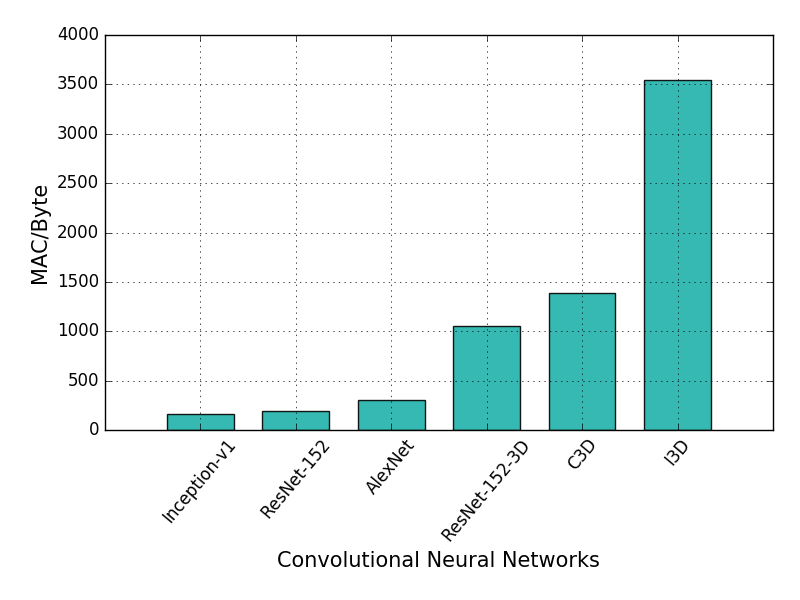}
    }
    \caption[]{Comparing popular 2D CNNs AlexNet~\cite{alexnet}, Inception~\cite{inception} and ResNet~\cite{resnet} with 3D CNNs C3D~\cite{c3d}, ResNet3D~\cite{resnet3d} and I3D~\cite{i3d}.}
    \label{fig:2dvs3d}
\end{figure*}

\subsection{Challenges in Accelerating 3D CNN Inference}

It is important to ask: given that 3D CNNs are a generalization of 2D CNNs, can a 2D CNN accelerator (e.g., Eyeriss~\cite{eyeriss}) efficiently evaluate a 3D CNN?
We find the answer is no, the key reasons being that the temporal dimension in 3D CNNs exposes new data reuse opportunities that cannot be captured by, and significantly exacerbates certain design issues present in, 2D CNN accelerators.

To the first point, contemporary 2D CNN accelerators are designed to exploit spatial (width and height) data reuse effectively\cite{eyeriss}. 
However, 3D CNNs feature data reuse opportunities in both spatial dimensions within a frame and the temporal dimension across frames. 
Without mechanisms to take advantage of temporal data reuse close to the arithmetic units, 2D accelerators must evaluate a 3D CNN ``frame by frame,'' which incurs significant memory system overheads.

To the second point, we make several important observations that collectively show how several design issues in 2D CNNs get significantly more pronounced when working with 3D CNNs.

\para{Observation 1: Working set size exceeds on-chip memory.}
Figure~\ref{fig:size_2dvs3d} shows the number of Bytes needed to store inputs (containing activations) and filters (containing weights)---two data types in CNNs---for both modern 2D and 3D CNNs.
(Partial sums, the third major data type, are not shown.)
Evident from the figure, 3D CNN memory footprint for both data types far exceeds the typical on-chip memory provisioned on energy-efficient accelerators.
Yet, 2D accelerators (e.g., \cite{ucnn,scnn}) typically pin a specific data type (e.g., inputs) statically in on-chip memory, expecting that data type to always fit.
This is sub-optimal in cases when the other data type (e.g., filters) fits in on-chip memory, in which case the correct strategy is to pin filters. 

\para{Observation 2: On-chip memory requirements vary dramatically.}
Again observed from Figure~\ref{fig:size_2dvs3d}, input and filter memory requirements vary significantly across layers in the case of 3D CNNs. 
Yet, 2D CNN hardware resources are typically provisioned for the worst case layer (e.g., the input memory in \cite{scnn}).
This design philosophy cannot be used when designing for 3D CNNs, as memory fragmentation overheads stemming from provisioning for the worst case will be exacerbated.


\para{Observation 3: On-chip energy is more pronounced, relative to off-chip energy.}
Figure~\ref{fig:compute_2dvs3d} shows the ratio of multiply-accumulate operations (MACCs) to memory footprint (sum of input and filter storage) for 2D and 3D CNNs. 
This shows that \emph{data reuse}---the number of computations done per Byte of data---is higher for 3D CNNs.
This not only makes 3D CNNs significantly more compute bound, but also reduces the ratio of energy spent accessing the off-chip memory vs. overall energy compared to 2D CNNs.
Whereas off-chip accesses often consume the majority of energy in 2D CNN acceleration~\cite{Cambriconx}, the increased reuse in 3D CNNs means the major factors governing on-chip energy---in particular efficiently managing buffers and reuse---become more prominent.

\subsection{This Paper}

This paper proposes Morph: a novel accelerator for accelerating 3D CNN inference.
Based on the above observations, the key design decision we made when architecting Morph was to maximize \emph{configuration-time flexibility}, which allows Morph hardware to adapt to each layer of each 3D CNN it targets.
Based on Observations 1 and 2, Morph hardware can tile any (or all) 3D CNN data types and share on-chip storage between tiles regardless of their size.
Further, Morph can change the order in which tiles are scheduled to processing resources, both in time and space (also called ``dataflow''~\cite{eyeriss}).
For example, when the input data footprint is small (see later layers in Figure~\ref{fig:size_2dvs3d}), it is prudent to pin inputs in on-chip memory and stream through the weights; vice versa for early layers.
Finally, based on Observation 3, Morph endows these degrees of flexibility to multiple levels of on-chip buffering, so as to maximize efficiency on-chip as well as off-chip.

To summarize, this paper makes the following contributions:

\begin{enumerate}
\item We design and implement Morph: a flexible 3D CNN hardware accelerator.  To our knowledge, Morph is the first ASIC accelerator targeting 3D CNNs.
\item We codesign a software framework that finds proper configuration-time parameters (tile sizes, loop order, loop parallelism) for each layer of each 3D CNN that runs on the Morph accelerator.
\item
We evaluate Morph across multiple 3D and 2D CNNs, comparing our final proposal to a less-flexible version of our accelerator and to Eyeriss~\cite{eyeriss}, a popular accelerator for 2D CNNs.
Evaluated on state-of-the-art 3D CNNs, Morph achieves an average $2.5\times$ reduction in energy consumption and an average  $5.1\times$ improvement in performance/watt compared to the less-flexible baseline.
Morph further achieves a $15.9\times$ average energy reduction on 3D CNNs when compared to Eyeriss.
Lastly, we implement Morph in RTL and synthesize the design in a 32~nm process, finding the cost of flexibility to be 5\% area over a similarly-provisioned inflexible accelerator.
\end{enumerate}

\section{Background}
\label{sec:background}

\subsection{Methods for Video Understanding}
\label{subsec:methods}

Broadly, there are two categories of works in video understanding: hand-crafted and learning-based methods. 
Hand-crafted methods evolved from their successful counterparts in image recognition, such as STIP\cite{stip} and SIFT\cite{sift}. 
Several early methods, e.g., \cite{hog}, generated histogram descriptors for spatio-temporal volumes to generate features. 
Subsequent methods used hand-crafted approaches such as improved Dense Trajectories\cite{idt} to further boost accuracy. 

\begin{figure}[!t]
  \begin{centering}
  \includegraphics[width=\columnwidth]{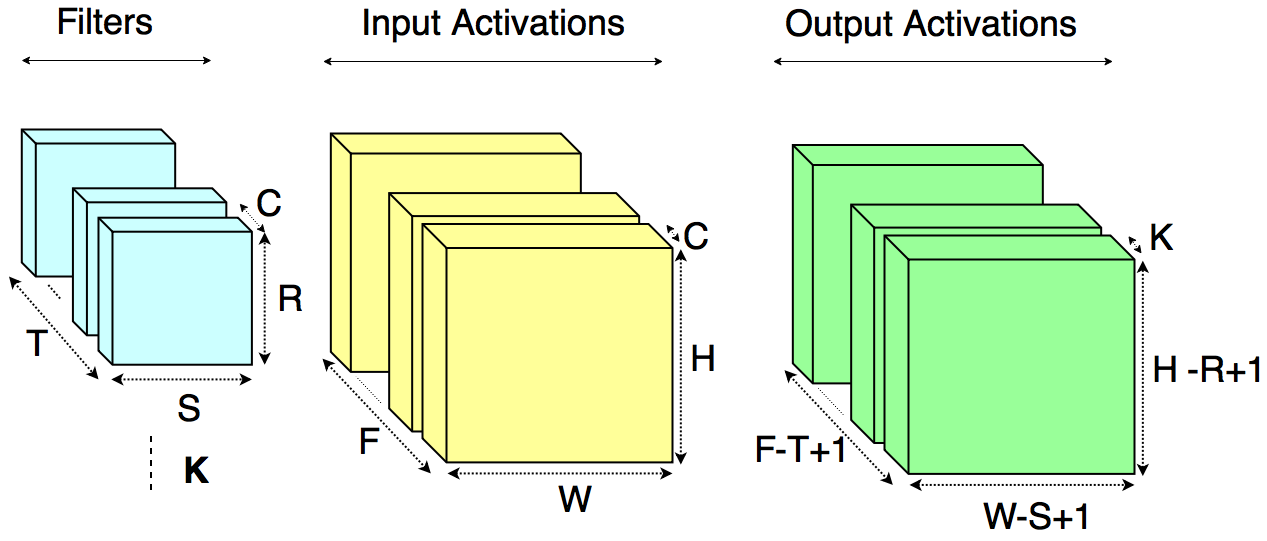}
  \caption{\label{fig:conv3D}
            \footnotesize 
            3D convolution operation on one input video.} 
  \end{centering}
\end{figure}

\subsection{3D Convolution}
\label{subsec:3dconv}

After the breakthrough in image recognition using convolutional neural networks (CNNs)~\cite{alexnet}, CNNs and other learning-based methods became the main approach for processing videos. 
Applying 2D convolutions and temporal pooling over videos was explored by \cite{karpathy}.
Using frame-based spatial and optical flow-based temporal streams of 2D convolutions was proposed by \cite{twostream}. 
\cite{conv3d, c3d} proposed 3D convolutions to model spatio-temporal features, which inspired significant follow-up work in using 3D convolution and its derivatives in video understanding\cite{i3d, p3d, t3d, r2p1d}. 
While there is no consensus on the best method for video understanding, 3D CNNs currently hold the state-of-the-art results on recent datasets.
Thus, our focus for the rest of the paper will be to accelerate 3D convolution.

In 3D convolution, filters (made up of weights) are moved spatially as well as temporally, performing dot products at each spatial-temporal position in the input.
Consider a video input of spatial resolution $H\times W$, $F$ frames (temporal) and $C$ channels. 
Then, if $K$ filters of spatial size $R\times S$, temporal size $T$ and $C$ channels are 3D-convolved with the input, it produces an output of spatial size $(H-R+1)\times(W-S+1)$ with $K$ channels and $(F-T+1)$ frames.
Note that $R \leq W$, $S \leq H$ and $T \leq F$.
This is graphically depicted in Figure~\ref{fig:conv3D} and in Algorithm~\ref{alg:conv3d}.

    

\textbf{Remark.}
2D convolution on images is a special case of 3D convolution with $F=1$ and $T=1$.
That is, hardware supporting 3D convolutions can also support 2D convolution.

\subsection{Compute Requirements in 3D Convolution}

In image recognition applications based on 2D CNNs, the 2D convolution kernel dominates the computation~\cite{2dconvruntime}.
Analysis of the compute requirements of state-of-the-art 3D CNNs~\cite{c3d, resnet3d, i3d} shows us that 3D convolution is even more compute bound, relative to 2D convolution. 
For example, during inference in C3D~\cite{c3d}, 3D convolution makes up over 99.8\% of compute (the remaining 0.2\% being video pre-processing, fully connected layers, Relu activations and pooling layers). 
Hence, it is important to accelerate and optimize 3D convolutions for efficient video understanding.

\textbf{Remark.} 
While a 3D CNN's model size grows on the order of $T$ over a similarly provisioned 2D CNN, the compute requirements increase on the order of $F * T$.
This means on-chip energy will play a larger role in total energy, relative to DRAM energy, compared to 2D CNNs. This effect is observed in Figure~\ref{fig:compute_2dvs3d}.


\begin{algorithm}[!t]
\caption{3D convolution operation. 
}\label{alg:conv3d}
\begin{algorithmic}[1]
\algrenewcommand{\algorithmiccomment}[1]{\hskip3em$\rightarrow$ #1}
\Procedure{conv3D}{$\mathrm{in}\;\mathbf{I},\mathbf{F}; \mathrm{out}\;\mathbf{O}; H,W,F,C,R,S,T$}
   \For{$k\gets0$ to $K$}
       \For{$f\gets0$ to $F-T+1$}
           \For{$w\gets0$ to $W-S+1$}
               \For{$h\gets0$ to $H-R+1$}
                   \State{$out = 0$} 
                   \For{$(r,s,c,t)\gets$ (0,0,0,0) to $(R,S,C,T)$} 
                   \State $out \mathrel{+}= \mathbf{I}$[f+t][c][w+s][h+r] * $\mathbf{F}$[t][c][s][r]
                \EndFor
               \State{$\mathbf{O}$[f][k][w][h] $\gets out$}
               \EndFor
           \EndFor
       \EndFor
   \EndFor
\State \textbf{return} $\mathbf{O}$
\EndProcedure
\end{algorithmic}
\end{algorithm}

\subsection{Tiling}
\label{subsec:tiling}

Accelerators access data through a hierarchy of memories ranging from expensive off-chip memory to relatively cheaper on-chip buffers. 
As observed from Figure~\ref{fig:size_2dvs3d}, inputs and filters in a 3D CNN will likely not fit in the accelerator's on-chip buffers. 
In such cases, the only (efficient) way to perform 3D convolution is to tile the data such that each tile fits on chip.

\begin{figure}[!t]
  \begin{centering}
  \includegraphics[width=0.98\columnwidth]{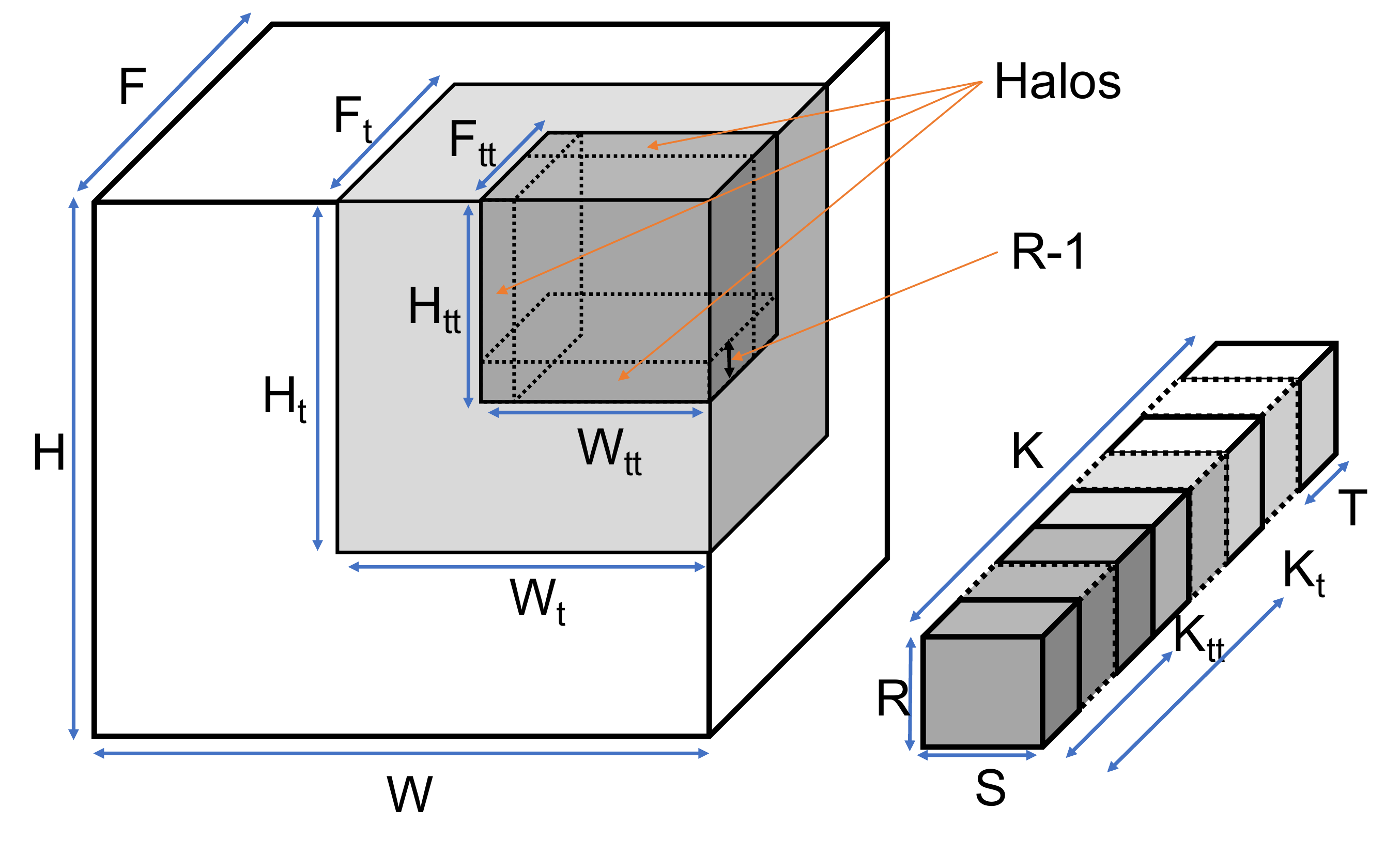}
  \caption{\label{fig:tiling}
            \footnotesize 
            Tiled 3D Convolution. Note that similar tiling can be performed in the $C$ dimension.
            Halo size is given for he $H$ dimension only, for a stride of 1.
            }
  \end{centering}
\end{figure}

To improve tiling effectiveness, on-chip buffers can further be organized in a hierarchy, down to registers close to the compute.
In general, it is possible to have as many levels of tiling as the number of memory hierarchies used.
Figure~\ref{fig:tiling} shows tiled 3D convolution with two levels of tiling.
We use $X_t$ to represent the first level tile of a parameter $X$, $X_{tt}$ to denote the second level tile (a \emph{sub-tile} of $X_t$), etc.


When the input does not fit on chip, it is broken into tiles of size $H_tW_tC_tF_t$ that do fit, as shown in Figure~\ref{fig:tiling}. 
When not all of $K$ filters fit in on-chip memory, they are broken into tiles of $K_t$ filters, where each of the $K_t$ filters have size $RSC_tT$.
Filters in the filter tile are then convolved over the input tile to complete computation for that tile.
Input and filter tiles can further form sub-tiles in lower level on-chip buffers, such as $H_{tt}W_{tt}C_{tt}F_{tt}$ and $K_{tt}$.
Note that we tile only in $K$, $F$, $W$, $H$ and $C$ dimensions. 
The remaining variables---$R$, $S$ and $T$---are generally small values that range between 1 and 11~\cite{c3d,i3d}, hence not considered for tiling.
The minimum tile size of inputs and filters is $RSC_tT$.

Since convolution involves a sliding window, adjacent input tiles overlap if the convolution stride is less than the tile dimensions. 
This is referred to as the data \emph{halo}~\cite{scnn}, and is depicted in Figure~\ref{fig:tiling}.
Note that in 2D convolution, there are halos in the $WH$ dimensions.
In 3D convolution, there are halos in the $WHF$ dimensions.

\subsection{Loop Order}
\label{subsec:looporder}


From Algorithm~\ref{alg:conv3d}, we see that the result of 3D convolution remains the same irrespective of the loop order, as the dot product operation is commutative.
Yet, different loop orders entail different on-chip memory requirements, data movement patterns, compute and memory resource utilizations, and halo overheads.
Interchanging loops has been widely studied as a means to improve performance~\cite{loopinterchange} and has a significant impact on 3D convolution efficiency because loop variable bounds vary significantly in modern 3D CNNs.

In section~\ref{subsec:tiling}, we saw there could be as many levels of tiling as the number of memory hierarchies. 
Similarly, for each level of tiling, we have a separate loop order which determines the order of dimensions in which the tiles are fetched.
We represent loop orders as lists $[XYZ]$, where $X$, $Y$ and $Z$ are dimensions, and $X$/$Z$ represent the outermost/innermost dimensions of the loop, respectively.
$X$ and $Z$ are also known as the \emph{minor} and \emph{major} traversal order, respectively.
For example, the $H_{tt}W_{tt}C_{tt}F_{tt}$ tile shown in Figure~\ref{fig:tiling} can be accessed in the order of any combination of spatial ($H$, $W$), channel ($C$) or temporal ($F$) dimensions inside the $H_tW_tC_tF_t$ tile.
The loop order $[WHCKF]$ moves through the $F$ spatial dimension first, $C$ second, etc.

\para{Data transfers.}
The loop order specifies when and how data transfers occur into and out of the current buffer level.
Consider any loop order containing dimensions $W$, $H$, $C$, $F$ and $K$.
Data transfers into the current level buffer occur at the following points:
\begin{itemize}
    \item \textbf{Filters.} The next tile of filters is loaded in the innermost loop labeled $C$ or $K$.
    \item \textbf{Inputs.} The next tile of inputs is loaded in the innermost loop labeled $W$, $H$, $C$ or $F$.
    \item \textbf{Partial sums.} The next tile of partial sums is loaded in the innermost loop labeled $W$, $H$, $K$ or $F$.
\end{itemize}
For example, given the loop order $[WHCKF]$, filter tiles are loaded in the second-to-innermost loop ($K$), inputs in the innermost loop ($F$), and partial sums in the innermost loop ($F$).

Data reuse is directly proportional to loop order.
In the above example, filters are reused spatially in the $F$ dimension.
Due to input tile halo (Section~\ref{subsec:tiling}), consecutive input tiles overlap. 
In this case, we take advantage of slide reuse and do not re-fetch the overlapped region in the major dimension.
For example, the above loop order does not re-fetch overlapped data in the $F$ dimension as it slides in the $F$ dimension, but does re-fetch overlapped data in other dimensions ($W$ and $H$) when it reaches the end of the $F$ dimension.

\subsection{Parallelizing Convolution}
\label{subsec:parallel}

For a given loop order, a hardware accelerator may choose to execute loop iterations in sequence or in parallel.
Parallelizing iterations entails spatially scheduling different iterations across processing elements (PEs).
For example, parallelizing in the $K$ dimension in Algorithm~\ref{alg:conv3d} conceptually converts the \code{for} loop in line 2 to a \code{parallel-for} loop.

Which dimension offers the largest opportunity for parallelism varies across layers as the size of each dimension varies over the layers. 
For example: it is easy to parallelize across inputs in early layers, but not in later layers as the input dimensions get smaller in later layers.
A fixed choice of dimension for parallelism, therefore, can hurt accelerator efficiency as observed by~\cite{flexiflow}.
Parallelism is well studied by prior works in 2D convolution~\cite{flexiflow}, and 3D convolution allows parallelization in the time dimension~($F_p$) in addition to the spatial~($H_p$, $W_p$) and filter ($K_p$) dimensions.

We note that loop order and PE parallelism, together, give the accelerator's \emph{dataflow}~\cite{eyeriss,scnn}.

\section{Motivation}
\label{sec:motivation}

\begin{figure*}[t]
    \centering
    \subfloat[Effect of outer loop order.\label{fig:motivation_off_chip}]{
    \includegraphics[width=0.305\textwidth]{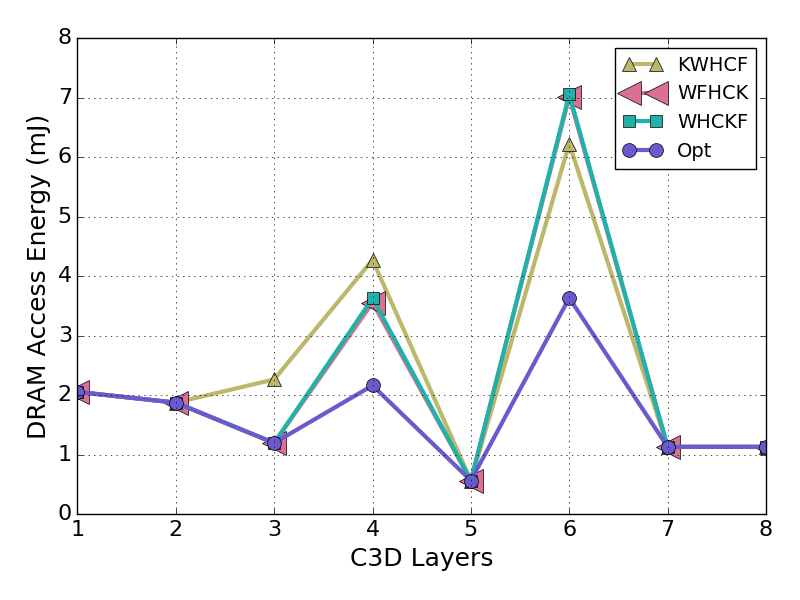}
    } ~
    \subfloat[L2 buffer allocation across layers for the Opt configuration in Figure~\ref{fig:motivation_off_chip}.
    \label{fig:motivation_tile_size}]{
    \includegraphics[width=0.3\textwidth]{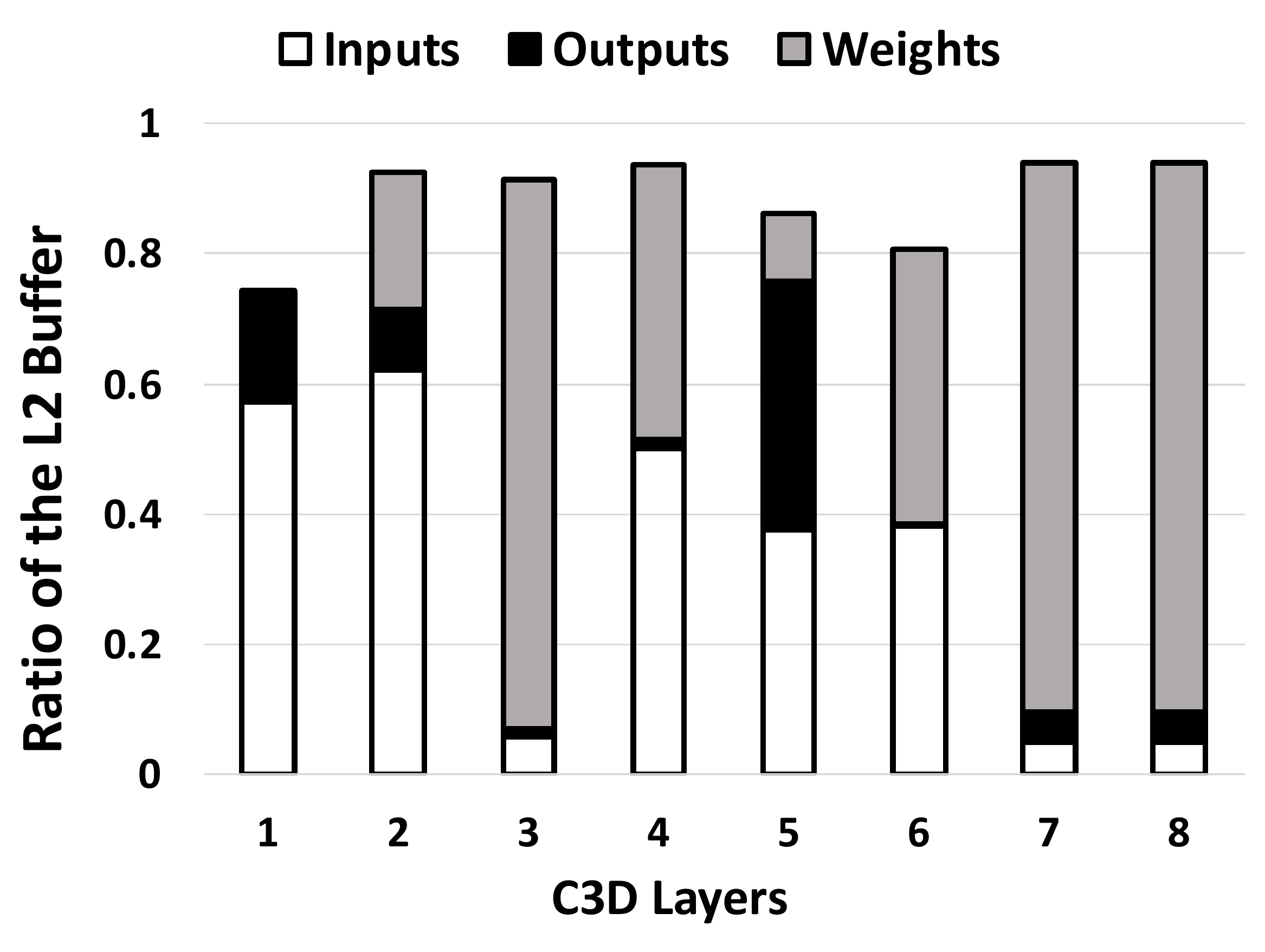}
    } ~
    \subfloat[Effect of inner loop order.\label{fig:motivation_on_chip}]{
    \includegraphics[width=0.305\textwidth]{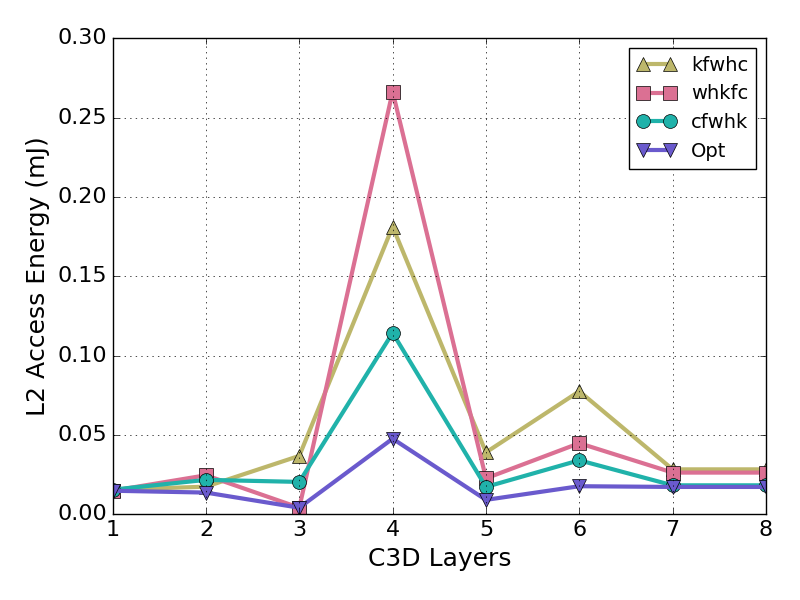}
    }
    
    \caption[]{ Access energy for different layers of C3D~\cite{c3d} for various loop orders and tile sizes. Figures denote outer loop orders with upper case letters and inner loop orders with lower case letters.
    }
    \label{fig:motivation}
\end{figure*}

In this section, we provide details on the benefits of configuration-time flexibility, in particular the impact of changing loop order, tile size and degree of PE parallelism per-layer in 3D CNN inference.
All experiments are shown for a representative 3D CNN, called C3D\cite{c3d}, and assume an accelerator with three levels of on-chip buffer which can be flexibly partitioned between inputs, filters and partial sums---similar to our final evaluated design in Section~\ref{sec:eval}.
Thus, changing tile size does not cause memory fragmentation.
On-chip buffer sizes are a 1~MB L2 (last level), 64~KB L1, and 16~KB L0, all of which are double buffered, which represent typical parameters for an edge accelerator. 
Buffer hierarchies are considered inclusive for the rest of the paper.

For simplicity, we break the loop order into two components for the rest of the paper:
\begin{enumerate}
\item \textbf{Outer loop order} refers to the order in which the input tile $H_tW_tC_tF_t$ and filter tile $K_tRSC_tT$ is fetched from off-chip memory to the last level of on-chip buffer (L2 in this case).
The possible loop order variables include spatial ($H$, $W$), channel ($C$), temporal ($F$) and filter ($K$).
\item \textbf{Inner loop order} refers to the order in which the input tile $H_{t..t}W_{t..t}C_{t..t}F_{t..t}$ and filter tile $K_{t..t}RSC_{t..t}T$ is accessed from a higher on-chip buffer to the lower level on-chip buffers or the compute elements.
The possible loop order variables include spatial ($h$, $w$), temporal ($f$), channel ($c$) and filter ($k$), denoted lower case to distinguish from the outer loop order.
We use the same inner loop order for scheduling tiles between L2-L1 and L1-L0 buffers.
\end{enumerate}

\textbf{Remark.}  We assume 8-bit inputs and weights for our experiments as this a standard across several works ~\cite{eie,tpu} for 2D CNN inference. To the best of our knowledge, 3D CNNs for video understanding have not been studied for precision, but we will assume that similar results for 2D would hold for 3D.



\subsection{DRAM Access Energy}
\label{subsec:dram_reads}

First, we analyze DRAM energy consumption across layers as a function of outer loop order.
Note that the DRAM access pattern is fully determined by outer loop order and tile sizes in the L2 memory.

We compare four configurations in Figure~\ref{fig:motivation_off_chip}: $[KWHCF]$, $[WFHCK]$ and $[WHCKF]$ and a best-case scenario ``Opt'' explained below.
As discussed in Section~\ref{sec:intro}, the frequency that weights are iterated through, which is determined by the position of $K$ in the outer loop order, is a first-order constraint.
Thus, we show outer loop orders $[KWHCF]$ and $[WFHCK]$ to illustrate extreme points, where the L2 buffer is weight or input stationary~\cite{eyeriss}, respectively.
We show outer loop order $[WHCKF]$ as it achieves the best energy overall, averaged across all layers.
Finally, whereas the above three configurations use the same outer loop order across all layers, Opt picks whichever outer loop order is optimal for each layer. 
For each bar in the figure, we sweep tile sizes and inner loop orders, and plot the configuration with lowest overall (off-chip plus on-chip) energy to isolate the effect of outer loop order.\footnote{Note that since this architecture flexibly shares buffering between data types, the sum of all L2 tile sizes is bounded by 512~KB.}

There are several important observations.
First, the energy cost for the $K$-extreme loop orders follows the observations in Section~\ref{sec:intro} and Figure~\ref{fig:size_2dvs3d}, where input/weight working set varies dramatically across layers.
Specifically, outer loop orders with $K$ in the inner loop do better in early layers, but perform worse in the later layers. 
This is because the filters progressively get larger in later layers, making it prudent to iterate through the  filters fewer times in those layers.
The trend is reversed for inputs and the spatial dimensions $W,H$, which get progressively smaller in later layers.
Second, the best loop order overall, $[WHCKF]$, does not provide the lowest energy for each layer individually, and all configurations incur large overheads relative to Opt.
This indicates that 3D CNN accelerators should flexibly support different 
outer loop orders at layer granularity.
As such, we design Morph with sufficient flexibility to achieve Opt.

\textbf{Remark.} In Figure~\ref{fig:motivation_off_chip}, layers 1, 2, 5, 7 and 8 have the same DRAM energy regardless of outer loop order.
This is because in those layers, one data type fits entirely in the L2 (see Figure~\ref{fig:size_2dvs3d}), meaning that the same tile appears multiple times in a row.
We assume the accelerator does not re-fetch redundant data in these cases.

Figure~\ref{fig:motivation_tile_size} gives more insight on how the (flexible) L2 buffer is partitioned between inputs, weights and outputs for the Opt case in Figure~\ref{fig:motivation_off_chip}. Inputs occupy a larger buffer percentage in early layers, whereas filters occupy more space in later layers. The best energy-performance can be obtained when one of the data types can be fully accommodated in the L2, as observed for Layer 3 in Figure~\ref{fig:motivation_tile_size}, which fully fits the filters by allocating less space to inputs. Similarly for layers 5, 7 and 8, fitting the outputs entirely is beneficial as they can be carried over as inputs to the next layer. 
Overall, it is clear that buffer allocation flexibility is important for accelerator efficiency.

\subsection{On-chip Memory Access Energy}
\label{subsec:buffer_read}


Beyond DRAM accesses, tile size and inner loop order for the on-chip buffers also have a significant impact on the overall inference energy.
Figure~\ref{fig:motivation_on_chip} repeats a similar methodology as in Section~\ref{subsec:dram_reads}, varying inner loop order instead of outer loop order.
As done in Section~\ref{subsec:dram_reads}, we choose three inner loop orders---$[kfwhc]$, $[whkfc]$ and $[cfwhk]$~(average best), and sweep other parameters such as tile size to yield the lowest energy point for each inner loop order.

The takeaway is that observations similar to those made in Section~\ref{subsec:dram_reads} are also valid for tiling decisions made between on-chip memories. 
We see that the best performing inner loop order varies from layer to layer and the average best inner loop order is not the best solution in each layer.
As before, Opt, which selects the optimal inner loop order per layer, is significantly better than other strategies.
Hence, 3D CNN accelerators need to support flexible inner loop orders to minimize the energy spent in on-chip buffers.

We do not show experiments for on-chip buffer tile size, but note they are analogous to the experiment in Section~\ref{subsec:dram_reads}.

\subsection{Performance via PE Utilization}

Beyond energy, accelerator performance is maximized when all PEs are doing work.
Several earlier works in 2D CNN acceleration have noted that static dataflows with good PE utilization in early layers leads to poor PE utilization in later layers, and vice versa~\cite{scnn}. 
Follow-on/concurrent work has shown how distributing tiles in either the filter or spatial dimensions, depending on the layer, can get around this problem and enable high PE utilization for all layers~\cite{flexiflow, maeri}.

We observe that the root cause of PE under-utilization in 2D CNN accelerators---that weight volume grows towards later layers and input volume shrinks---also holds for 3D CNNs, as Figure~\ref{fig:size_2dvs3d} and previous analysis in this section have shown.
Thus, a 3D CNN accelerator which can flexibly parallelize tiles in different dimensions, as discussed in Section~\ref{subsec:parallel}, can likewise avoid PE utilization issues.


\subsection{Take-away}

This section has shown how an accelerator that performs best ``on average, across layers'' is sub-optimal relative to an accelerator that can adapt to the needs of each layer.
It is clear that there is no one optimal inflexible hardware configuration to efficiently run every layer of a 3D CNN, let alone across different 3D CNNs.

Accordingly, the rest of the paper develops a hardware architecture that can be configured to evaluate different loop orders, tile sizes and degrees of PE parallelism for each layer of inference.
We further develop a software infrastructure that pre-analyzes each layer so as to determine the optimal hardware configuration for each layer.
Together, these components enable highly efficient inference across a range of 3D CNNs with varying storage and compute requirements.

\section{Hardware Architecture}
\label{sec:arch}


\begin{figure}[!t]
  \begin{centering}
  \includegraphics[width=0.8\columnwidth]{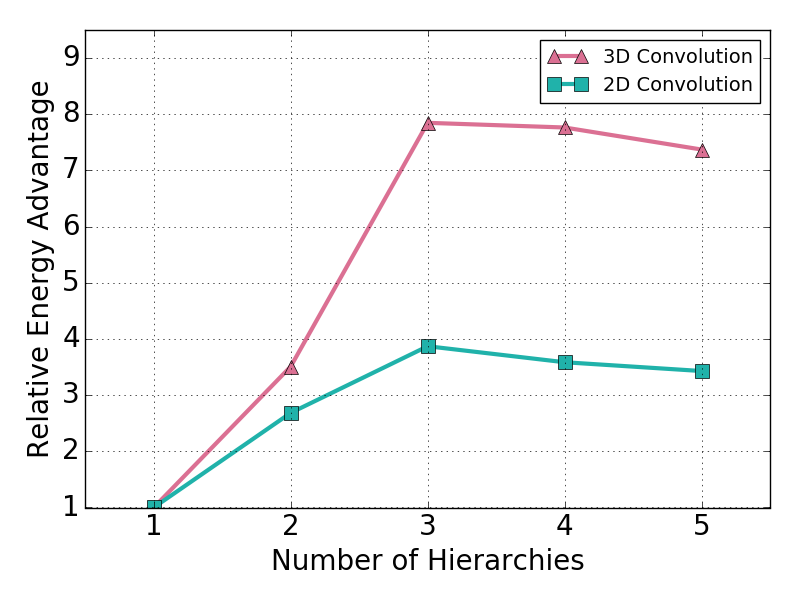}
  \caption{\label{fig:hierarchy_2dvs3d}
            \footnotesize 
            Sweep study for relative energy advantage for multi-level buffer hierarchies over a single level buffer hierarchy, for 3D and 2D convolution. 
            The graph assumes an input of size $112 \times 112 \times 3$~($HWC$) with $16$ frames ($F$) convolved with a filter of size $3 \times 3 \times 3$~($RSC$) with a temporal depth of $3$~($T$). Note that 2D convolution sets $F$ and $T$ to $1$.
            }
  \end{centering}
\end{figure}

In this section, we first describe design principles for an inflexible Morph base architecture. 
Then we show how modifications to the base architecture enable flexible tiling, loop ordering and PE parallelism.

\subsection{Base Architecture}
\label{subsec:arch_baseline}

3D CNNs are computationally similar to 2D CNNs, but with additional data reuse opportunities in the time dimension.
Filters in 2D CNNs slide in the spatial plane, along the $H$ and $W$ dimensions.
As a result, each input is reused in $R\times S$ dot products per filter, ignoring edge effects and assuming stride 1.\footnote{2D and 3D CNNs also offer input reuse in the filter $K$ dimension, as each of the $K$ filters works over the same spatial or spatial-temporal input.  Since this factor is the same in both 2D and 3D CNNs, we focus the discussion on slide reuse (in the $WH$ or $WHF$ dimensions), not filter reuse ($K$).}
To exploit this reuse, 2D CNN accelerators (e.g., \cite{eyeriss}) architect custom logic to re-read inputs in buffers close to the processing elements, without re-loading those inputs from higher level buffers.

3D CNNs generalize convolution to spatial-temporal dimensions, sliding in $W$, $H$ and $F$ dimensions.
This increases input reuse to a factor of $R\times S\times T$.
Running a 3D CNN on a stock 2D CNN accelerator results in sub-optimal efficiency due to the lack of support for temporal reuse.
A 2D CNN accelerator must perform 2D convolution on each of $T$ frames separately and then merge the resultant partially computed frame to generate a final output frame. This process repeats $F-T+1$ times to produce the complete output.  
This introduces large overhead in the form of on/off-chip buffer transfers per frame.
To mitigate this overhead, we design our accelerator to exploit the spatial-temporal reuse in on-chip buffers, e.g., close to the PEs, in an analogous fashion as 2D CNNs exploit spatial reuse.

\subsubsection{Buffer hierarchy}

A key pre-requisite to exploit data reuse is to design a deep enough on-chip buffer hierarchy to support all degrees of temporal locality in the data access pattern.
Thus, to start, we perform an analytic design space search to determine a sufficient number of levels of buffer hierarchy in Figure~\ref{fig:hierarchy_2dvs3d}.
In this experiment, for each buffer hierarchy (one level, two levels, etc.) we sweep possible loop orders and tile sizes, fixing the physical buffer size to the tile size, to isolate the effect of levels of hierarchy.
As in Section~\ref{sec:motivation}, we show whichever configuration yields lowest energy consumption.
The figure shows results for a representative layer.

Both 2D and 3D CNNs benefit from at least three levels of on-chip buffer hierarchy and, accordingly, our designs have a three-level memory hierarchy throughout the paper.
Interestingly, the effect of additional memory hierarchy in 3D CNNs is more pronounced: a three-level hierarchy yields a $7.8\times$ improvement over a one-level hierarchy, relative to a $3.8\times$ improvement for 2D CNNs.
The reason is due to additional halo effects present in 3D CNNs (Section~\ref{subsec:tiling}).
To prevent halo overhead, 3D CNNs prefer larger tiles/buffer sizes.
But, larger tiles lead to higher energy cost per access.
Adding another level of memory hierarchy brings down the access energy cost, distributing the data locality across the hierarchy.

We note that energy efficiency drops beyond three levels of buffer.
The reason is that reuse has already been sufficiently captured in three levels, and adding additional levels simply adds buffer reads/writes to buffer levels that cannot provide additional data reuse.  

\begin{figure}[!t]
  \begin{centering}
  \includegraphics[width=\columnwidth]{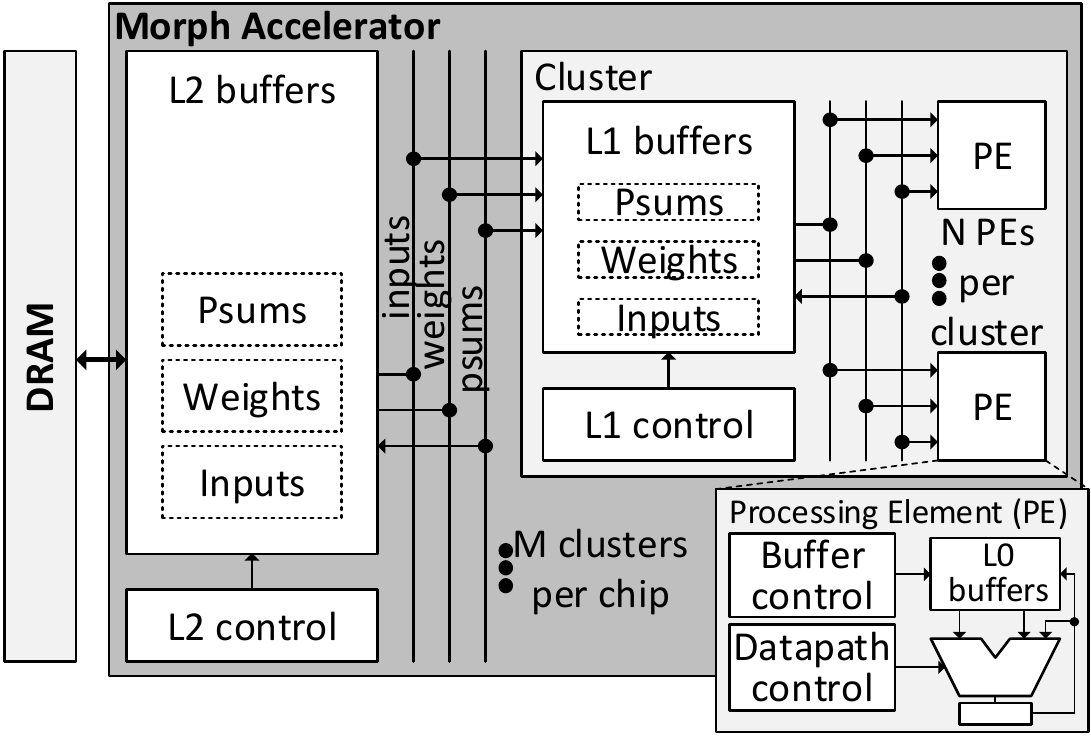}
  \caption{\label{fig:arch}
            \footnotesize 
            Top level view of the Morph base architecture.
            The final Morph architecture (Section~\ref{subsec:arch_morph}) adds logic to make buffers and control logic configurable to different loop orders, tile sizes and degrees of PE parallelism.} 
  \end{centering}
\end{figure}

\subsubsection{Architecture}

With the three-level buffer hierarchy in mind, Figure~\ref{fig:arch} shows the top level view of the Morph base architecture.
Morph base is built with $M$ compute clusters where each cluster consists of $N$ processing elements (PEs).
The accelerator consists of L2, L1 and L0 on-chip buffers, each statically partitioned to store inputs, filters and partial sums (psums).\footnote{As with 2D CNNs, psums are wider in bitwidth than input activations and weights.}
Each partition's size is pre-determined based on what is needed to support the worst-case tile size across target 3D CNNs (see 3D CNNs in Section~\ref{sec:cnn_eval}).
L0 buffers reside in the PEs; L1s in each cluster; and the L2 is the last-level buffer before DRAM.
Each buffer has static control logic, e.g., hard-coded FSMs, that governs where/when data is read/written, how many MACCs to perform, etc.
Inputs, filters and psums are transferred over three broadcast networks that connect the L2 to the L1s/clusters.
Each cluster has a separate (local) set of three broadcast networks which connect its L1 to local L0s/PEs.
Data is consumed by ALUs in each PE.
Each ALU supports $V_w$ lanes of vector multiply and add/accumulate (MACC) operations.
Vector lanes are provisioned across output channels (in the $K$ dimension).
Each PE has $V_w$ accumulator registers, one per lane, to reduce psum traffic between the L0 and ALU.
Finally, to remove dead time between processing tiles, all buffers are logically double buffered.

\subsubsection{Dataflow}
\label{sec:morph_base_dataflow}

The Morph base architecture implements a fixed loop order and tile size which we found to give the best average performance/watt across a range of 3D CNNs.
This methodology is analogous to that used for inflexible accelerators running 2D CNN inference (e.g., \cite{scnn,eyeriss}), and we elaborate on this process further in Section~\ref{sec:software}.

\begin{enumerate}
    \item \textbf{Outer loop order:} 
    Morph base implements an outer loop order of $[WHCKF]$. 
    This implies that the input tile is fetched to the L2 from DRAM in the $F$ dimension first, and repeats this traversal for all filters ($K$) before completing the other spatial dimensions.
    \item \textbf{Inner loop order:} 
    All levels of on-chip buffer in Morph base use an inner loop order of $[cfwhk]$. 
    The PEs go through all the filters in $K_{t..t}$ before sliding in spatial and temporal dimensions; maximizing the input reuse. 
    Finally, the process repeats for the next $C_{t..t}$ channels, reusing psums from the last iteration.
\end{enumerate}

Finally, Morph base parallelizes work across PEs using a fixed $H_p$ and $K_p$.


\subsubsection{On-chip networks}
\label{sec:impl:broadcast_noc}

All networks on chip (NoCs) in the Morph base architecture are simple broadcast networks, which can implement unicast-, multicast- and broadcast-style data transfers using a mask to indicate destination(s).


We argue that the large degree of data reuse present in 3D CNNs allows architects to build such simple NoCs without starving the compute units.
Consider the system shown in Figure~\ref{fig:arch} with $M\times N$ PEs using 1~Byte input activations and weights.
If each PE were to consume and finish using one unique input per cycle, the bus between L2 and L1 would need to transfer $M\times N$ bytes of input per cycle.
Because each input is reused $R \times S \times T$ times (stride 1), however, the bus between the L2 and L1s need only carry $\frac{ M \times N }{R \times S \times T}$ inputs per cycle to rate match the PEs in the steady state.
Note that this argument implies that rate matching is strictly easier for 3D CNN accelerators relative to 2D CNNs, as 3D CNNs have an additional factor $T$ reuse for inputs.

A similar argument allows us to reduce bus bandwidth between each L1 and its $N$ PEs. 
Required bus bandwidth for weights and psums is even less significant, as each weight and psum is reused $(W-R+1)\times (H-S+1)\times (F-T+1)$ and $C\times R \times S \times T$ times, respectively, which is larger than the factor $R\times S\times T$ reuse in each input.

Given the concrete dataflow from Section~\ref{sec:morph_base_dataflow}, we can set broadcast bus bandwidth based on expected 3D CNN parameters and desired compute throughput.
For example, given filters with $R=S=T=3$ and stride 1 (which is typical) and a desired compute throughput of $N*M=36*6=216$ MACCs/cycle ($N=36$ PEs for each of $M=6$ clusters) in the steady state, we only require a 64~bit bus between the L2 and L1s and a 32~bit bus between each L1 and its L0s to rate match the PEs.

\subsection{Morph: A Flexible Architecture}
\label{subsec:arch_morph}

We now propose changes to the Morph base architecture so that it can better match the needs of different 3D CNN layers.
In general, there are four aspects to imparting flexibility to rigid CNN accelerators, namely configurable buffers, PE control logic, NoCs~\cite{maeri} and datapaths~\cite{flexiflow, dna}.
We will add support for flexible buffer partitions to enable different tile sizes per 3D CNN data type without introducing buffer fragmentation.
We will add flexible control logic to enable different outer and inner loop orders.
Finally, we make minor modifications to the Morph base NoC and PE datapath to enable different degrees of PE parallelism.

Additional flexibility adds area, energy and frequency overhead.
We note that previous works have shown that the area and power consumed by the on-chip buffers in CNN accelerators dominates the control logic, ALUs and datapath.
Hence, despite the logic added to support flexibility, the overall area and power overhead is not significant (see Section~\ref{sec:eval}).




\subsubsection{Configurable Buffers}
\label{subsubsec:config_buf}

\begin{figure}
  \begin{centering}
  \includegraphics[width=\columnwidth]{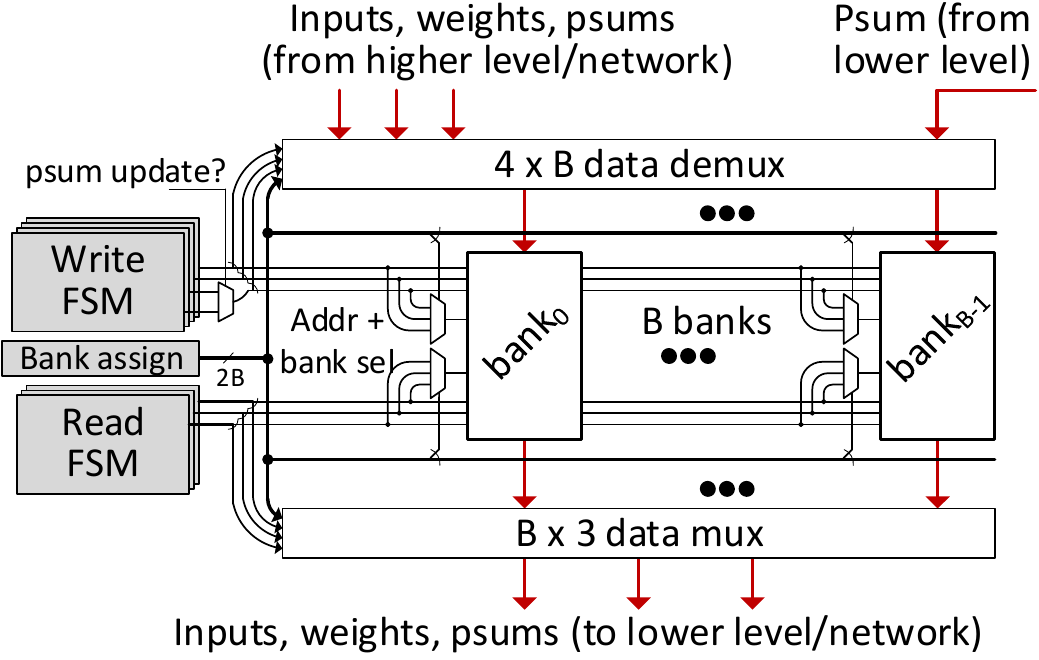}
  \caption{\label{fig:buffer}
            \footnotesize 
            Configurable buffer with $B$ banks.
            Shaded blocks are programmable.
            The bank assign logic outputs a  $2B$-bit wide vector that indicates, for each bank, whether that bank is assigned to inputs, filters or psums.
            }
  \end{centering}
\end{figure}

The goal of providing buffer configurability is to allow different tile sizes of inputs, filters and psums at each level of the buffer hierarchy, while minimizing internal fragmentation in each physical buffer. 
Figure~\ref{fig:buffer} shows the configurable buffer design we use in Morph.
This design is reused for each level of on-chip memory, and is used to share space between inputs, filters and psums within a level.

Each buffer is first divided into $B_i$ banks (for buffers in the L$i$-th level memory).
Each bank supports a single read and single write port.
For configuration purposes, $B_i$ for $i=0,1,2$ is exposed to software, which allocates memory at bank granularity across inputs, filters and psums (Section~\ref{sec:software}).
Banks are allocated to each data type contiguously, and base registers (``Bank assign'' in Figure~\ref{fig:buffer}) configured at layer start time denote the range of banks used to store each data type.
Parallel mux/demux logic is used to index into and read/write data into/out of each group of banks.
The output (read) mux is replicated for each of the three types of outputs (inputs, filters, psums) and since each data type reads one word per access, there are no bank conflicts.

Programmable FSMs (Section~\ref{subsubsec:control_logic}) are used to generate address patterns into each group of banks.
High-order address bits, along with the bank assignment registers, determine which bank is responsible for each type of read.
The full address is sent to each bank to derive a local bank address as well as bank select signals.
Thus, reading each data type activates only one bank to save energy.


This design is simple, however may lead to some internal fragmentation depending on how well tile sizes cleanly divide into banks. 
Further, banking in general runs the risk of increased area due to the use of less dense SRAM arrays.
In our evaluation, L2, L1 and L0 memories are decomposed into 16 banks, which was sufficient to support variable tile sizes efficiently.
In this regime, we found area overheads due to banking to be minimal.
For example, breaking a 1~MB L2 into 16 banks only added a $4.9$\% area overhead~\cite{cacti}.

\para{Buffer data width.}
Note that psums are wider than input activations and weights.
Specifically, given $P$-bit precision per activation/weight, a psum requires $2*P+\log_2(RSTC)$~bits to avoid overflow, due to the number of MACCs per dot product (Section~\ref{subsec:3dconv}). 
We handle this disparity in different ways for different buffering levels.
At the L1 and L2, we use wider word widths (i.e., $>P$), sufficient to read out one psum or more than one activation/weight per access.
This design has an added benefit that energy/bit/read decreases with wider SRAM word widths.
At the L0, we set the buffer word width to $P$~bits.
While this is appropriate for activations/weights,
psums require multiple cycles to access.
However, this design doesn't seriously degrade performance as psum reads/writes out of the L0 are infrequent relative to MACCs into the local accumulator register.


\para{Access priority.}
As done with Morph base, the buffer is logically double-buffered to avoid dead time between tiles.
All data types are written from higher level buffers over the broadcast network, and psums may be written back from lower levels.
When writes from a higher level and a psum update from a lower level happen in the same cycle, priority is given to the higher level and standard pipeline backpressure is used to stall the psum update.
We note that psum updates are relatively infrequent, because the accumulator register below the ALU filters psum writebacks to higher levels.

\subsubsection{Control Logic}
\label{subsubsec:control_logic}
The Morph base architecture uses fixed function control FSMs that implement logic for a specific loop order at each level of the memory hierarchy (Section~\ref{sec:morph_base_dataflow}).
These FSMs generate addresses into the data buffers, count how many MACCs to perform, when to read/write psums relative to performing MACCs, when processing all tiles is complete, etc.
Such control signals change significantly for different tile sizes and loop orders.
For example, when the loop order changes, loop bounds and memory access pattern into each tile changes, along with the frequency of events such as loading and unloading tiles between memory levels.


\begin{figure}
  \begin{centering}
  \includegraphics[width=\columnwidth]{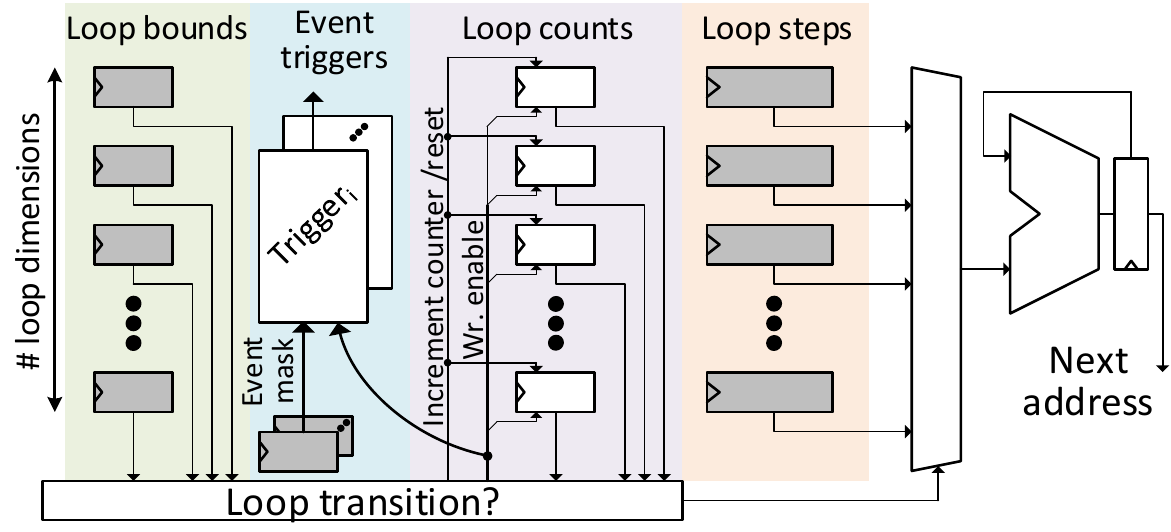}
  \caption{\label{fig:fsm}
            \footnotesize 
            Programmable read/write FSM responsible for generating addresses into buffers and other control signals such as tile done, etc.
            Shaded registers are configurable at layer start time.
            }
  \end{centering}
\end{figure}

To enable control flexibility for the final architecture, we use the configurable FSM shown in Figure~\ref{fig:fsm}.
The FSM is programmed by setting two sets of configurable registers which denote loop bounds and loop steps, for a parameterizable number of loops which is determined at design time.
The FSM walks through the loop using the loop bounds and accumulates a step into an output register.
That is, for a $D$-level loop, the user specifies bounds $b_0,\dots,b_{D-1}$ and steps $s_0,\dots,s_{D-1}$.
Each FSM `state' corresponds to an iteration of this $D$-level loop, given by iteration indices $i_0,\dots,i_{D-1}$ where $i_j<b_j$ for $j=[0,D-1]$.
That is, iteration indices behave like their software counterparts.
When entering each state, the FSM outputs the current value in the output register and one of the steps $s_j$ is added to that register for the next iteration (similar to \cite{ganax,bitfusion}).
Here, $j$ equals which loop is currently terminating (i.e., if all $i_k=b_k-1$ for $k=0,\dots,j\le D-1$) or 0 if no loop is terminating.


By setting loop bounds and loop steps appropriately, the accumulator gives the different address sequences needed by the buffers for different loop orders.
We also add logic to the FSMs, called triggers, to derive non-address control signals.
We observe that these events---e.g., end of tile, unloading/reloading a psum from the ALU accumulator register---occur at loop iteration boundaries. 
Thus, simple two-level logic with a programmable mask (``Event mask'' in the figure) can derive these signals from loop count reset signals already generated in the FSM. 

\subsubsection{Controlling PE Parallelism}

Finally, to support flexible degrees of PE parallelism we architect a mask register to control unicast, multicast and broadcast on the bus-based NoCs.
Within a layer, the degree of PE parallelism is fixed.
The exception is in the last round of tiles, which may occupy less PEs due to edge effects.
We handle this effect with a counter to track when the last round begins, and a second mask register which reconfigures the NoC for that round.

\section{Software Optimizations}
\label{sec:software}

%
%
%


In this section, we describe a software optimization framework that pre-analyzes 3D CNNs and finds the optimal tiling and loop order parameters per layer, using knowledge of the underlying Morph architecture. 
Popular software libraries such as TensorFlow~\cite{tensorflow}, Caffe~\cite{caffe} and Theano~\cite{theano} provide a high-level programming interface to users while masking the low-level optimizations specific to hardware. 
The software optimizer covered here would fit in as a library that gets called when 3D convolution is requested by the user and the chosen device is Morph.
We note that these optimizations need only be performed once per CNN.
After best-fit parameters are found once, a configuration file can be saved and recalled instead of re-running the analysis. 

The flow takes two inputs: 
(1) layer parameters which include sizes of input activations, weights and other parameters like stride (e.g., $W$, $H$, etc.); and 
(2) architecture details of Morph which include the number of PEs ($N$) and clusters ($M$), and the size of each L0, L1 and L2 memory with their banked configuration ($B_i$ for each $i$). 
The optimizer returns several ``best'' configurations (e.g., best performance, best performance/watt, etc.) which the user can select between. 
Each configuration specifies tile sizes, loop orders and spatial PE parallelism. 

\subsection{Generating Configurations}
\label{subsec:gen_config}

Based on (1) layer parameters and (2) architectural details, the optimizer first enumerates all possible configurations. 
First a parameter list is generated that includes:
\begin{itemize}
    \item All possible inner and outer loop orders.
    \item All possible last-level buffer (L2) tile sizes: $H_t$, $W_t$, $C_t$, $K_t$ and $F_t$.
    \item Parallelization parameters, such as $H_p$, $W_p$ and $K_p$.
\end{itemize}

Chosen L2 tile sizes serve as a starting point for later heuristics that select sub-tile sizes for remaining buffer levels.
To reduce search time, the L2 tile size and degree of PE parallelism search space can be discretized.



The optimizer takes the cartesian product of the parameter list to enumerate the configurations, where each configuration is [outer loop order, inner loop order, $H_t$, $W_t$, $C_t$, $K_t$, $F_t$, $H_p$, $W_p$, $K_p$].
Once generated, the following steps process each configuration:


\subsection{Generating Metadata}
\label{subsec:metadata}

Based on the current configuration, this step generates metadata required for further calculations.
This includes: the number of iterations the chosen tile has to perform to complete the convolution, storage requirements for each of the tiles, overlapped regions for tile slides~(halos\cite{scnn}), the final output size, etc.

\subsection{Memory Allocation}
\label{subsec:mem_alloc}

In this step, the optimizer uses a heuristic to set the tile size for each data type (inputs, weights, psums) for each level of on-chip memory below the last level, given each starting configuration defined in Section~\ref{subsec:gen_config}.

Consider an $N$ level on-chip memory hierarchy where each level buffer has size $L_n$. 
Let $T_n$ represent a list giving the tile size for each data type (input, weights, psums). 
The \emph{allocate} heuristic finds a $T_n$ for the $n^{th}$ level buffer such that:
    \begin{itemize}
    \item $T_{min} \leq T_n \leq T_{n+1}$ where $T_{min}$ represents the minimum tile size (for each data type) required to perform a 3D convolution.
    That is, sub-tiles are smaller than tiles, sub-sub-tiles smaller than sub-tiles, etc.
    \item $f_{reuse}(T_n,$ inner loop order, buffer size, $B_n)$ is maximized, where $f_{reuse}$ is a function that returns the amount of reuse for each tile, for a given inner loop order and buffer (size and num banks).
    \item $\sum T_n \leq L_n$, i.e., the sum of the tile sizes does not exceed the physical buffer size.
\end{itemize}

Given tile candidates and configuration information, $f_{reuse}$ calculates the ratio of buffer fills (from a higher level buffer) to reads and updates (from lower levels).
For input tiles, reuse comes from sliding (which reduces halo cost) and output channels (as the same input is used for each output channel).
For filter tiles, reuse comes spatially as the same filter is used for different spatial positions in the input.

\emph{allocate} is called level by level, starting from level $N-1$ and going down to 0 (as level $N$ is specified in the configuration), until all tile sizes are specified.
For each level, allocate searches a small percentage of the overall tile size space as follows.
For a $D$-dimensional tile, allocate generates $2^D$ tile sizes where each tile size corresponds to setting each dimension in the tile to have minimum or maximum size.
For example, a two-dimensional tile with dimensions $XY$ has four sizes, corresponding to setting each dimension to max or min: (max, max), (max, min), (min, max), (min, min).
Consider a tile in level $n+1$ which has dimension $X_t$, meaning it has dimension $X_{tt}$ in level $n$.
Then the max size for dimension $X$ occurs when $X_t=X_{tt}$. The minimum size follows the rules in Section~\ref{subsec:tiling}.
Once three sets of tile sizes are generated for the three data types, we take the cartesian product to generate candidates for $T_n$, and each of these are tested using $f_{reuse}$.

\subsection{Performance and Power Calculation}
\label{subsec:rule_gen}

Once memory allocation is complete, there is enough information to compute the number of operations performed in each PE, the number of PEs active at a given time (PE utilization) and the number of reads/writes that occur in each buffer.
We use a linear energy model to convert the number of reads/writes/operations to the expected energy consumed for the layer.
We use an analytic model to convert PE utilization and other configuration metadata (such as the number of tiles computed per PE, the number of reads per tile, etc.) to accelerator wall clock time. 
\subsection{Final Configuration Generation}
\label{subsec:final_res}

The optimizer performs the above steps for each configuration and produces corresponding power and performance results.
Once all results are available, it is straightforward to optimize for power or performance or performance/power, etc.

The final configuration can then be used to derive all state needed to configure Morph from Section~\ref{subsec:arch_morph}, e.g., bank assignments and FSM state.


\section{Evaluation}
\label{sec:eval}

\subsection{Measurement Setup}

We evaluate Morph using a whole-chip performance and energy model, and design/synthesize the Morph PEs with RTL written in Verilog.
All designs are evaluated in a 32~nm process with a 1~GHz clock.
For the energy model, energy numbers for arithmetic units are taken from~\cite{marktalk}, scaled to 32~nm.
SRAM energies are taken from CACTI~\cite{cacti}.
For all SRAMs, we assume \texttt{itrs-lop} as this decreases energy per access, but still yields SRAMs that meet timing at 1~GHz.
DRAM energy is counted at 20~pJ/bit~\cite{marktalk}.
Network on chip (NoC) energy is extrapolated based on the number and estimated length of wires in the design (using our PE area and L2 SRAM area estimates from CACTI).
We assume the NoC uses low-swing wires~\cite{cacti-lowswing}, which are low power, however consume energy each cycle (regardless of whether data is transferred) via differential signaling.

\begin{figure*}[!t]
  \begin{centering}
  \includegraphics[width=\textwidth]{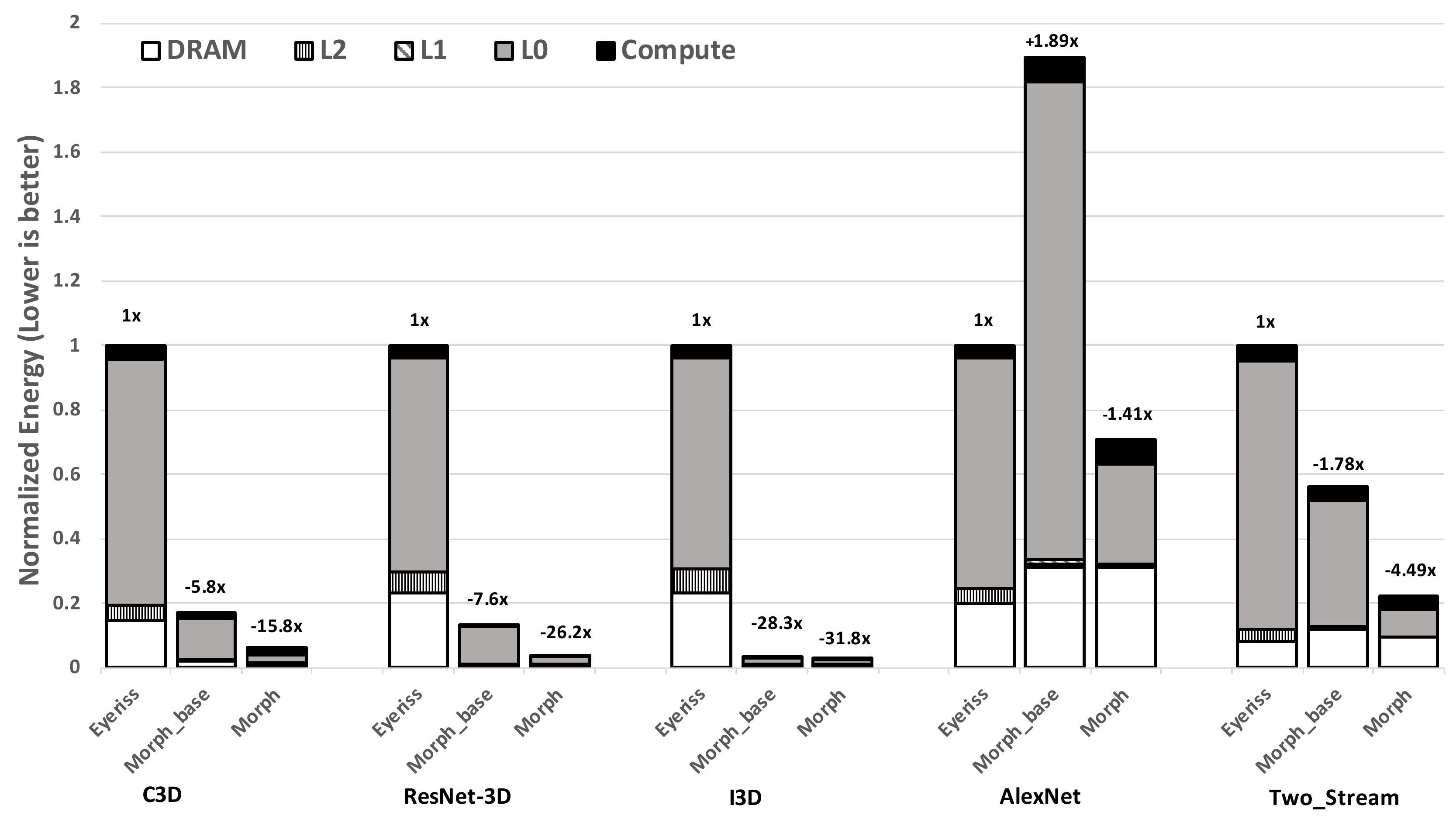}
  \caption{\label{fig:power_consumption}
            \footnotesize 
             Energy consumption for various state-of-the-art 2D and 3D CNNs when run on Eyeriss, Morph and Morph\_base. Values are normalized to Eyeriss.
             }
     \label{fig:energy_fig}
  \end{centering}
   
\end{figure*}

\subsection{Points of Comparison}
\label{sec:comparison_points}

We evaluate the design against three variants: 
\begin{enumerate}
    \item \textbf{Morph}: 
    Includes mechanisms endowing hardware flexibility (Section~\ref{subsec:arch_morph}) and uses the Morph software analysis to determine best loop order and buffer partitions.
    
    \item \textbf{Morph\_base}: A baseline system that runs an average best loop order generated by the Morph optimizer (Section~\ref{sec:software}), specifically, outer loop order $[WHCKF]$ and inner loop order $[cfwhk]$.
    Table \ref{tab:buffer_partition} shows the static partitioning for the on-chip buffers.
    We choose the partition sizes that give the average best energy efficiency across all DNNs under test.
    
    \begin{table}[h]
\centering
\caption{On-chip buffer partitions.}
\label{tab:buffer_partition}
\begin{tabular}{c|c|c|c}
    \hline
    \hline
    \textbf{Hierarchy} &      \textbf{Inputs}      & \textbf{Outputs} & \textbf{Weights}\\
    \hline
    \hline        
    L2       &   38.5\% & 40\% & 21.5\%\\ 
    \hline
    L1       &    40\% & 10\% & 50\% \\
    \hline 
    L0      &     40\% & 10\% & 50\% \\
    \hline
\end{tabular}
\end{table}
    
   
    \item \textbf{Eyeriss}:
    To compare against a well-optimized 2D CNN accelerator, we simulate Eyeriss\cite{eyeriss} using the nnflow simulator~\cite{tetris}. 
    We take $100\%$ density for both input and filters and normalize all the parameters in Eyeriss with Morph in terms of maximum compute power and available on-chip memory as shown in Table~\ref{tab:morph_config}. 
    Eyeriss evaluates a 3D CNN ``frame by frame'' as described in Section~\ref{sec:arch}. 
\end{enumerate}

We use 32~nm-based energy numbers for both buffer access and compute across all configurations for a fair comparison. 
Morph and Morph\_base use three levels of on-chip buffer as described in Section~\ref{sec:arch}, and the L2, L1 and L0 are sub-divided into 16 banks each.


\begin{table}[h]
\centering
\caption{Simulation parameters.}
\label{tab:morph_config}
\begin{tabular}{c|c|c}
    \hline
    \hline
    \textbf{Parameters} &      \textbf{Morph}      & \textbf{Eyeriss} \\
    \hline
    \hline
    PEs         & 16 (per cluster)   & $24\times 32$ \\
    \hline
    Clusters    &     6         & $-$ \\
    \hline
    Vector Width &     8          & 1 \\
    \hline        
    L2 Size      &      1024 kB   & 1408 kB \\ 
    \hline
    L1 Size      &     64 kB (per cluster)     & $-$ \\
    \hline 
    L0 Size     &      16 kB (per PE)      & 2 kB (per PE) \\
    \hline
\end{tabular}
\end{table}

\subsection{CNNs Evaluated}
\label{sec:cnn_eval}

We evaluate the proposed methods on the following networks:
\begin{enumerate}
    \item \textbf{C3D}~\cite{c3d}: Owing to its popularity and wide adoption in action recognition.
    \item \textbf{I3D}~\cite{i3d}: As it currently holds the state-of-the art results on the Kinetics\cite{activitynet} video dataset.
    \item \textbf{3D ResNet-50}~\cite{resnet3d}: A 3D version of the popular ResNet-50\cite{resnet}.
    \item \textbf{2-Stream}~\cite{twostream}: A 2D network that runs on multiple input frames.
     \item \textbf{AlexNet}\cite{alexnet}: One of the earliest and most popular 2D-CNNs for image recognition.
\end{enumerate}


\subsection{Energy Analysis}

\begin{table*}[!h]
\centering
\caption{C3D configuration optimized for energy by the Morph software analysis.}
\label{tab:kernels}
\begin{tabular}{|c|c|c|c|c|c|c|c|}
    \hline
    \multirow{2}{*}{Layer} & \multirow{2}{*}{Outer Loop Order} & \multirow{2}{*}{Inner Loop Order} & \multicolumn{4}{c|}{Configuration}\\\cline{4-7} 
    &&&
    $K_t$ & $H_t$ &  $F_t$ & $K_p*V_w$\\
    \hline
    layer1 & $KWFHC$ & $cwhfk$ & $64$ & $114$ &  $16$ & $8$ \\
    \hline
    layer2 & $KWHCF$ & $cfwhk$ & $128$ & $30$ &  $16$ & $8$ \\
    \hline
    layer3a & $KWHCF$ & $kcfwh$ & $16$ & $28$ & $8$ & $8$ \\
    \hline
    layer3b & $WFKHC$ & $whckf$ & $8$ & $16$ &  $6$ & $8$ \\
    \hline
    layer4a & $WFKHC$ & $whckf$ & $8$ & $14$ &  $4$ & $8$ \\
    \hline
    layer4b & $WHCKF$ & $kcfwh$ & $16$ & $14$ &  $4$ & $16$\\
    \hline
    layer5a & $WFKHC$ & $whckf$ & $32$ & $7$ & $2$ & $16$ \\
    \hline
    layer5b & $WFKHC$ & $whckf$ & $32$ & $7$ &  $2$ & $16$ \\
    \hline
\end{tabular}
\end{table*}



Figure~\ref{fig:energy_fig} shows the energy consumption of I3D, C3D, 3D ResNet-50, 2-Stream and AlexNet for Morph compared to the Morph\_base and Eyeriss.
A 3D architecture adaptive to the varying memory requirements in 3D CNNs can exploit more data reuse for different layers compared to an accelerator with a fixed strategy.
Effective data reuse at all levels of memory hierarchy leads to fewer accesses to the higher level buffers and expensive off-chip memory, thus saving energy.
This effect can be seen in Figure~\ref{fig:energy_fig}, where Morph shows $2.5\times$ on average improvement in energy over Morph\_base. 

Both Morph\_base and Morph significantly outperform Eyeriss~\cite{eyeriss} in terms of energy consumption on 3D CNNs.
This is mainly because Eyeriss cannot exploit temporal data reuse nor can it choose loop orders for different 3D CNN layers.
As the number of frames increase, the efficiency gap widens; as evident from I3D which uses 64 frames compared to 16 frames in C3D.
This is due to the increased temporal data reuse opportunity with more frames, which both Morph designs can exploit.
Interestingly, Eyeriss outperforms Morph\_base on AlexNet.
This is because the Morph\_base design'sL0 buffer is provisioned for 3D CNNs, where a larger tile helps extract proportionally more data reuse.
However, 2D CNNs like AlexNet do not benefit from larger tile sizes---the additional energy cost per byte outweighs the improvements in reuse. Morph, however, slightly outperforms Eyeriss on AlexNet, owing to the improved tiling parameters and flexible loop ordering.

To show further insights, each bar in Figure~\ref{fig:energy_fig} is broken down into five components: DRAM access energy, L2 global buffer, L1 cluster buffer, L0 and compute. 
The results prove our claim in Section~\ref{sec:intro} that the increased compute to memory ratio in 3D CNNs results in large on-chip energy consumption.
From Figure~\ref{fig:energy_fig}, we see that Morph is effectively able to reduce the DRAM, L2, L1 and L0 energy components by $1.6\times$, $3.2\times$, $2.5\times$ and $4\times$, respectively, on average relative to Morph\_base.
This demonstrates the effectiveness of flexibly configuring loop order at each level of the hierarchy and allowing flexibility in buffer partitions and tile sizes.

Table~\ref{tab:kernels} shows the results from the Morph software analysis for C3D. 
The table lists the configurations chosen for each layer of C3D, tuned for minimal energy consumption when run on the Morph architecture. 
The first two columns show the outer and inner loop orders and the rest show the tiling parameters.
This reflects how the loop orders and tile sizes vary across the layers to improve energy efficiency.

\subsection{Performance-per-watt Analysis}

\begin{figure}[!t]
  \begin{centering}
  \includegraphics[width=\columnwidth]{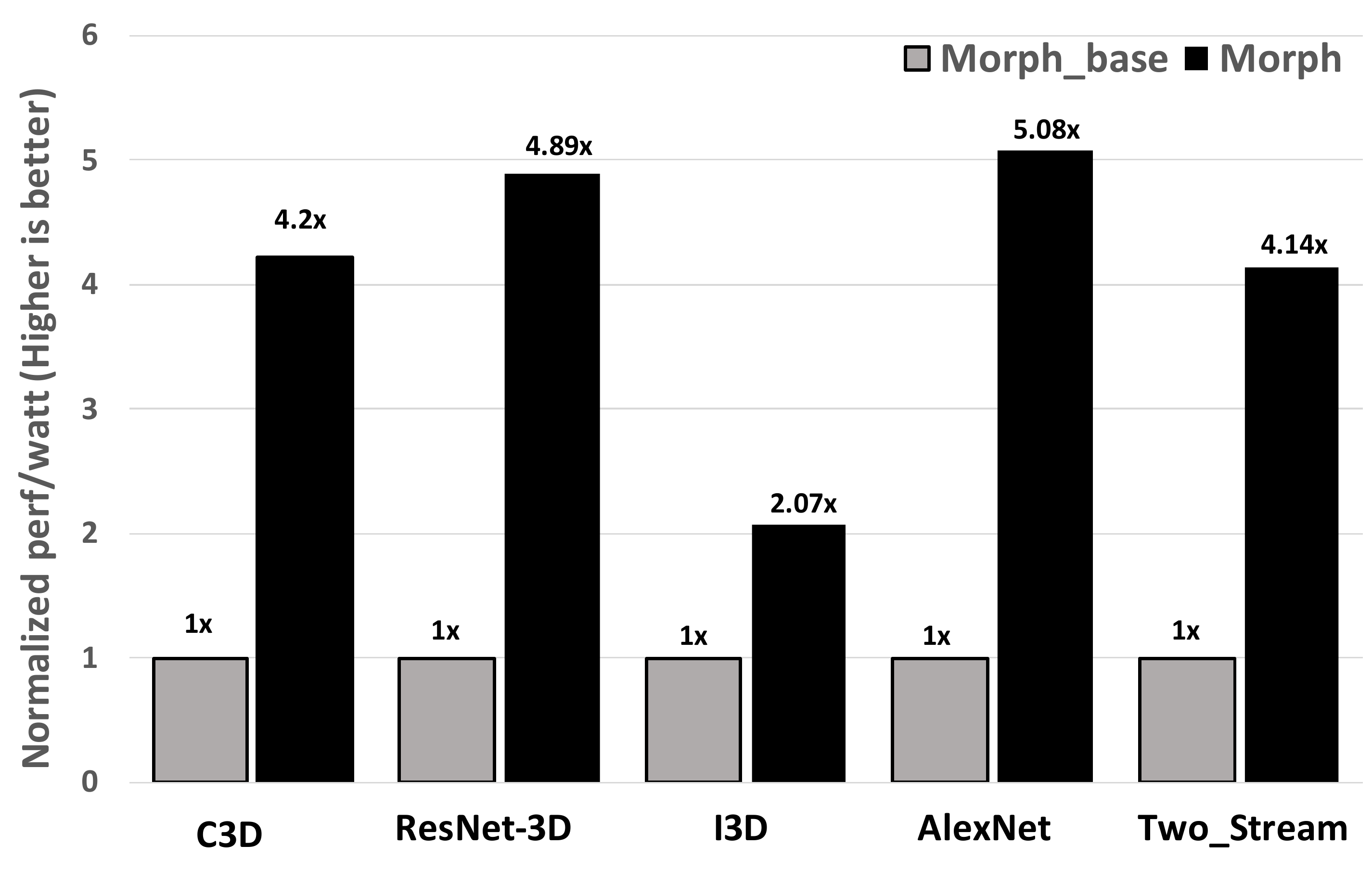}
  \caption{\label{fig:power_consumption}
            \footnotesize 
             Performance/watt comparison between Morph and Morph\_base.
             }
    \label{fig:ppw_fig}
  \end{centering}
\end{figure}

Given that both Morph and Morph\_base have the same theoretical maximum GFLOPs, any performance improvement should come from improved PE utilization. 
Figure~\ref{fig:ppw_fig} shows the performance-per-watt characteristics of Morph normalized to Morph\_base. 
Morph delivers $4\times$ on average improvement over Morph\_base; this can be attributed to the improved PE utilization achieved by adaptive loop orders and parallelization.
Morph adaptively chooses spatial parallelization parameters like $H_p$ and $W_p$, in and across layers.
This helps to keep the utilization of compute resources high in edge cases such as when tile size is not an integer multiple of the dimension size.
$K_p$ provides parallelization across filters, which can help keep the PEs busy, even when the inputs get smaller such as in later layers. 
2D accelerators like \cite{scnn} suffer from PE utilization challenges in the later layers due to the diminishing input size, and Morph's performance/watt improvement of $5.08\times$ in AlexNet~(2D-CNN) suggests that Morph adapts well to this problem and chooses adequate parallelization to improve overall performance.

\subsection{Hardware Implementation}
Finally, Table~\ref{tab:pe_area_breakdown} shows the area overhead of Morph mechanisms at the PE. 
We implemented both Morph and Morph\_base PEs in Verilog, using 8~bit precision weights/activations.
Synthesis used a 32~nm commercial process, and both designs met timing at 1~GHz.  
Area numbers for SRAM were obtained from CACTI~\cite{cacti} and the area for logic comes from synthesizing in a 32~nm process technology.

\begin{table}[h]
\centering
  \caption{
  Morph PE area breakdown (in $mm^2$).
  }
{\footnotesize
\centering
  \begin{tabular}{l | c | c | c}
    \hline 
    \hline
    Component               & Morph\_base & Morph & \% change\\
    \hline \hline
    L0 buffer               &0.041132      &0.042036 & 2.19\% \\ 
    Arithmetic              &0.00306      &0.00366 & 19.36\% \\
    Control Logic           &0.00107      &0.00182 & 70.59\% \\
    \hline
    Total                   &0.04526     &0.04751  & 4.98\% \\
    \hline 
  \end{tabular}
}
  \label{tab:pe_area_breakdown}
\end{table} 

Morph divides L0 into 16 banks, thus adding additional area overhead compared to statically partitioned monolithic SRAMs in Morph\_base. Arithmetic added a few changes to support flexible loop orders, which increases the area by 19\%. As expected, control logic sees a relative large increase in area (over 70\%), which is due to the increased complexity of the read/write FSMs and the control logic added for buffer partitioning. 
However, the overall area increase is almost negligible at 5\%. 
This should not come as a surprise, given the large area occupied by the on-chip memories.

\section{Related Work}
\label{sec:related}

\textbf{CNN accelerators.}
Due to the recent popularity of 2D CNNs in image recognition and related tasks, there have been a plethora of works proposing new architectures to accelerate 2D CNNs~\cite{diannao,dadiannao,shidiannao, eyeriss}.
There have been no works that accelerate 3D CNNs in ASICs, although we note several recent works\cite{3dcnn,fc3d} which have explored hardware acceleration of 3D CNNs on FPGAs.
On one hand, FPGAs are reconfigurable and thus can directly adapt to different CNN configurations.
On the other hand, FPGAs pay for this reconfigurability with reduced compute density and clock speeds.
The Morph architecture introduces several points of flexibility which allows it to adapt well to different CNNs, yet retain the efficiency benefits of an ASIC.
Lastly, a recent trend has been to exploit sparsity in inputs and weights, to save compute and compress the model\cite{scnn,eie,cnvlutin,cambricon,ucnn}. 
We do not study sparsity in 3D CNNs and consider it a future work, but expect that many of the same ideas from sparse 2D CNN accelerators~\cite{eie,Cambriconx,cnvlutin,scnn,ucnn} to apply.

\textbf{Adaptive accelerators.}
There has been a recent interest in the community towards flexible accelerators for 2D CNNs. FlexiFlow~\cite{flexiflow} proposes flexibility in choosing the dimension of parallelization for better PE utilization, but keeps the loop orders static. DNA~\cite{dna} proposes to reconfigure the datapaths to support different dataflows. MAERI~\cite{maeri} proposes using a reconfigurable NoC to support different dataflows. All the above mentioned works, however, only target 2D-CNNs. In comparison, Morph offers higher degrees of flexibility in terms of tile sizes, loop orders and parallelism achieved with flexible buffering and programmable control logic.

\textbf{Design space search for CNN accelerators.}
For ASICs like Morph, only the software component can vary to support different types of CNNs. 
Several prior works study how to search for optimal hardware configurations for FPGA-based accelerators~\cite{cnnopt,suda,looporder}, where the entire design can be re-parameterized at configuration time.
These designs employ techniques such as loop unrolling and loop interchange, similar to Morph.
As stated above, a disadvantage is their reliance on FPGAs, which incurs an area and performance hit due to the FPGA fabric.



\section{Conclusion}
\label{sec:conclusion}

This paper proposed Morph, a novel 3D CNN accelerator designed for video understanding, which can adaptively support different tile sizes and loop orders. 
We also proposed the Morph optimizer, which determines efficient hardware parameter settings for each layer of each target 3D CNN.  
Putting it all together, our accelerator significantly outperforms an inflexible but similarly provisioned baseline accelerator, as well as the Eyeriss 2D CNN accelerator, on a set of state-of-the-art 3D CNNs.
We view our work as a first step towards enabling real-time video understanding on edge devices and advocate imparting flexibility in CNN accelerators to improve power and performance efficiency.


\bibliographystyle{ieeetr}

\begin{thebibliography}{10}

\bibitem{alexnet}
A.~Krizhevsky, I.~Sutskever, and G.~E. Hinton, ``Imagenet classification with
  deep convolutional neural networks,'' in {\em NIPS'12}.

\bibitem{inception}
C.~Szegedy, W.~Liu, Y.~Jia, P.~Sermanet, S.~Reed, D.~Anguelov, D.~Erhan,
  V.~Vanhoucke, A.~Rabinovich, {\em et~al.}, ``Going deeper with
  convolutions,'' in {\em CVPR'15}.

\bibitem{resnet}
K.~He, X.~Zhang, S.~Ren, and J.~Sun, ``Deep residual learning for image
  recognition,'' in {\em CVPR'16}.

\bibitem{vgg}
K.~Simonyan and A.~Zisserman, ``Very deep convolutional networks for
  large-scale image recognition,'' {\em ArXiv'14}.

\bibitem{mnist}
Y.~LeCun, C.~Cortes, and C.~Burges, ``Mnist handwritten digit database,'' {\em
  AT\&T Labs [Online]. Available: http://yann. lecun. com/exdb/mnist}, vol.~2,
  2010.

\bibitem{cifar10}
A.~Krizhevsky and G.~Hinton, ``Learning multiple layers of features from tiny
  images,'' tech. rep., Citeseer, 2009.

\bibitem{imagenet}
J.~Deng, W.~Dong, R.~Socher, L.-J. Li, K.~Li, and L.~Fei-Fei, ``Imagenet: A
  large-scale hierarchical image database,'' in {\em CVPR'09}.

\bibitem{eyeriss}
Y.-H. Chen, J.~Emer, and V.~Sze, ``Eyeriss: A spatial architecture for
  energy-efficient dataflow for convolutional neural networks,'' in {\em ACM
  SIGARCH Computer Architecture News}, vol.~44, pp.~367--379, IEEE Press, 2016.

\bibitem{cnvlutin}
J.~Albericio, P.~Judd, T.~Hetherington, T.~Aamodt, N.~E. Jerger, and
  A.~Moshovos, ``Cnvlutin: Ineffectual-neuron-free deep neural network
  computing,'' in {\em ACM SIGARCH Computer Architecture News}, vol.~44,
  pp.~1--13, IEEE Press, 2016.

\bibitem{dadiannao}
Y.~Chen, T.~Luo, S.~Liu, S.~Zhang, L.~He, J.~Wang, L.~Li, T.~Chen, Z.~Xu,
  N.~Sun, {\em et~al.}, ``Dadiannao: A machine-learning supercomputer,'' in
  {\em MICRO'14}.

\bibitem{shidiannao}
Z.~Du, R.~Fasthuber, T.~Chen, P.~Ienne, L.~Li, T.~Luo, X.~Feng, Y.~Chen, and
  O.~Temam, ``Shidiannao: Shifting vision processing closer to the sensor,'' in
  {\em ACM SIGARCH Computer Architecture News}, vol.~43, pp.~92--104, ACM,
  2015.

\bibitem{scnn}
A.~Parashar, M.~Rhu, A.~Mukkara, A.~Puglielli, R.~Venkatesan, B.~Khailany,
  J.~Emer, S.~W. Keckler, and W.~J. Dally, ``Scnn: An accelerator for
  compressed-sparse convolutional neural networks,'' in {\em ISCA'17}.

\bibitem{ucnn}
K.~Hegde, J.~Yu, R.~Agrawal, M.~Yan, M.~Pellauer, and C.~Fletcher, ``Ucnn:
  Exploiting computational reuse in deep neural networks via weight
  repetition,'' in {\em ISCA'18}.

\bibitem{twostream}
K.~Simonyan and A.~Zisserman, ``Two-stream convolutional networks for action
  recognition in videos,'' in {\em NIPS'14}.

\bibitem{karpathy}
A.~Karpathy, G.~Toderici, S.~Shetty, T.~Leung, R.~Sukthankar, and L.~Fei-Fei,
  ``Large-scale video classification with convolutional neural networks,'' in
  {\em CVPR'14}.

\bibitem{c3d}
D.~Tran, L.~Bourdev, R.~Fergus, L.~Torresani, and M.~Paluri, ``Learning
  spatiotemporal features with 3d convolutional networks,'' in {\em ICCV'15}.

\bibitem{i3d}
J.~Carreira and A.~Zisserman, ``Quo vadis, action recognition? a new model and
  the kinetics dataset,'' in {\em CVPR'17}.

\bibitem{conv3d}
S.~Ji, W.~Xu, M.~Yang, and K.~Yu, ``3d convolutional neural networks for human
  action recognition,'' {\em TPAMI'13}.

\bibitem{ucf101}
K.~Soomro, A.~R. Zamir, and M.~Shah, ``Ucf101: A dataset of 101 human actions
  classes from videos in the wild,'' {\em ArXiv'12}.

\bibitem{activitynet}
F.~C. Heilbron, V.~Escorcia, B.~Ghanem, and J.~C. Niebles, ``Activitynet: A
  large-scale video benchmark for human activity understanding,'' in {\em
  CVPR'15}.

\bibitem{moments}
M.~Monfort, B.~Zhou, S.~A. Bargal, A.~Andonian, T.~Yan, K.~Ramakrishnan,
  L.~Brown, Q.~Fan, D.~Gutfruend, C.~Vondrick, {\em et~al.}, ``Moments in time
  dataset: one million videos for event understanding,'' {\em ArXiv'18}.

\bibitem{youtube8m}
S.~Abu-El-Haija, N.~Kothari, J.~Lee, P.~Natsev, G.~Toderici, B.~Varadarajan,
  and S.~Vijayanarasimhan, ``Youtube-8m: A large-scale video classification
  benchmark,'' {\em ArXiv'16}.

\bibitem{cisco}
``Cisco visual networking index: Global mobile data traffic forecast update,
  2016-2021 white paper,''

\bibitem{resnet3d}
K.~Hara, H.~Kataoka, and Y.~Satoh, ``Can spatiotemporal 3d cnns retrace the
  history of 2d cnns and imagenet?,'' {\em ArXiv'17}.

\bibitem{Cambriconx}
S.~Zhang, Z.~Du, L.~Zhang, H.~Lan, S.~Liu, L.~Li, Q.~Guo, T.~Chen, and Y.~Chen,
  ``Cambricon-x: An accelerator for sparse neural networks,'' MICRO'16.

\bibitem{stip}
I.~Laptev, ``On space-time interest points,'' {\em IJCV'05}.

\bibitem{sift}
P.~Scovanner, S.~Ali, and M.~Shah, ``A 3-dimensional sift descriptor and its
  application to action recognition,'' in {\em MM'07}.

\bibitem{hog}
I.~Laptev, M.~Marszalek, C.~Schmid, and B.~Rozenfeld, ``Learning realistic
  human actions from movies,'' in {\em CVPR'08}.

\bibitem{idt}
H.~Wang and C.~Schmid, ``Action recognition with improved trajectories,'' in
  {\em ICCV'13}.

\bibitem{p3d}
Z.~Qiu, T.~Yao, and T.~Mei, ``Learning spatio-temporal representation with
  pseudo-3d residual networks,'' in {\em ICCV'17}.

\bibitem{t3d}
G.~Varol, I.~Laptev, and C.~Schmid, ``Long-term temporal convolutions for
  action recognition,'' {\em TPAMI'17}.

\bibitem{r2p1d}
D.~Tran, H.~Wang, L.~Torresani, J.~Ray, Y.~LeCun, and M.~Paluri, ``A closer
  look at spatiotemporal convolutions for action recognition,'' {\em ArXiv'17}.

\bibitem{2dconvruntime}
J.~Cong and B.~Xiao, ``Minimizing computation in convolutional neural
  networks,'' in {\em ICANN'14}.

\bibitem{loopinterchange}
M.~E. Wolf and M.~S. Lam, ``A data locality optimizing algorithm,'' in {\em ACM
  Sigplan Notices}, vol.~26, pp.~30--44, ACM, 1991.

\bibitem{flexiflow}
W.~Lu, G.~Yan, J.~Li, S.~Gong, Y.~Han, and X.~Li, ``Flexflow: A flexible
  dataflow accelerator architecture for convolutional neural networks,'' in
  {\em HPCA'17}.

\bibitem{eie}
S.~Han, X.~Liu, H.~Mao, J.~Pu, A.~Pedram, M.~A. Horowitz, and W.~J. Dally,
  ``Eie: efficient inference engine on compressed deep neural network,'' in
  {\em ISCA'16}.

\bibitem{tpu}
N.~P. Jouppi, C.~Young, N.~Patil, D.~Patterson, G.~Agrawal, R.~Bajwa, S.~Bates,
  S.~Bhatia, N.~Boden, A.~Borchers, R.~Boyle, P.-l. Cantin, C.~Chao, C.~Clark,
  J.~Coriell, M.~Daley, M.~Dau, J.~Dean, B.~Gelb, T.~V. Ghaemmaghami,
  R.~Gottipati, W.~Gulland, R.~Hagmann, C.~R. Ho, D.~Hogberg, J.~Hu, R.~Hundt,
  D.~Hurt, J.~Ibarz, A.~Jaffey, A.~Jaworski, A.~Kaplan, H.~Khaitan,
  D.~Killebrew, A.~Koch, N.~Kumar, S.~Lacy, J.~Laudon, J.~Law, D.~Le, C.~Leary,
  Z.~Liu, K.~Lucke, A.~Lundin, G.~MacKean, A.~Maggiore, M.~Mahony, K.~Miller,
  R.~Nagarajan, R.~Narayanaswami, R.~Ni, K.~Nix, T.~Norrie, M.~Omernick,
  N.~Penukonda, A.~Phelps, J.~Ross, M.~Ross, A.~Salek, E.~Samadiani, C.~Severn,
  G.~Sizikov, M.~Snelham, J.~Souter, D.~Steinberg, A.~Swing, M.~Tan,
  G.~Thorson, B.~Tian, H.~Toma, E.~Tuttle, V.~Vasudevan, R.~Walter, W.~Wang,
  E.~Wilcox, and D.~H. Yoon, ``In-datacenter performance analysis of a tensor
  processing unit,'' in {\em ISCA '17}.

\bibitem{maeri}
H.~Kwon, A.~Samajdar, and T.~Krishna, ``Maeri: Enabling flexible dataflow
  mapping over dnn accelerators via reconfigurable interconnects,'' in {\em
  ASPLOS'18}.

\bibitem{dna}
F.~Tu, S.~Yin, P.~Ouyang, S.~Tang, L.~Liu, and S.~Wei, ``Deep convolutional
  neural network architecture with reconfigurable computation patterns,'' {\em
  VLSI'17}.

\bibitem{cacti}
N.~Muralimanohar and R.~Balasubramonian, ``Cacti 6.0: A tool to understand
  large caches,'' 2009.

\bibitem{ganax}
A.~Yazdanbakhsh, H.~Falahati, P.~J. Wolfe, K.~Samadi, N.~S. Kim, and
  H.~Esmaeilzadeh, ``Ganax: A unified mimd-simd acceleration for generative
  adversarial networks,'' in {\em ISCA'18}.

\bibitem{bitfusion}
H.~Sharma, J.~Park, N.~Suda, L.~Lai, B.~Chau, V.~Chandra, and H.~Esmaeilzadeh,
  ``Bit fusion: Bit-level dynamically composable architecture for accelerating
  deep neural network,'' in {\em ISCA'18}.

\bibitem{tensorflow}
M.~Abadi, P.~Barham, J.~Chen, Z.~Chen, A.~Davis, J.~Dean, M.~Devin,
  S.~Ghemawat, G.~Irving, M.~Isard, {\em et~al.}, ``Tensorflow: A system for
  large-scale machine learning.,'' in {\em OSDI ’16}.

\bibitem{caffe}
Y.~Jia, E.~Shelhamer, J.~Donahue, S.~Karayev, J.~Long, R.~Girshick,
  S.~Guadarrama, and T.~Darrell, ``Caffe: Convolutional architecture for fast
  feature embedding,'' in {\em MM'14}.

\bibitem{theano}
F.~Bastien, P.~Lamblin, R.~Pascanu, J.~Bergstra, I.~Goodfellow, A.~Bergeron,
  N.~Bouchard, D.~Warde-Farley, and Y.~Bengio, ``Theano: new features and speed
  improvements,'' {\em ArXiv'12}.

\bibitem{marktalk}
M.~Horowitz, ``Computing's energy problem (and what we can do about it).''
  ISSCC, 2014.

\bibitem{cacti-lowswing}
A.~N. Udipi, N.~Muralimanohar, and R.~Balasubramonian in {\em HiPC'09}.

\bibitem{tetris}
M.~Gao, J.~Pu, X.~Yang, M.~Horowitz, and C.~Kozyrakis, ``Tetris: Scalable and
  efficient neural network acceleration with 3d memory,'' {\em ACM SIGOPS
  Operating Systems Review}, vol.~51, no.~2, pp.~751--764, 2017.

\bibitem{diannao}
T.~Chen, Z.~Du, N.~Sun, J.~Wang, C.~Wu, Y.~Chen, and O.~Temam, ``Diannao: A
  small-footprint high-throughput accelerator for ubiquitous
  machine-learning,'' {\em ACM Sigplan Notices}, vol.~49, no.~4, pp.~269--284,
  2014.

\bibitem{3dcnn}
J.~Shen, Y.~Huang, Z.~Wang, Y.~Qiao, M.~Wen, and C.~Zhang, ``Towards a uniform
  template-based architecture for accelerating 2d and 3d cnns on fpga,'' in
  {\em FPGA'18}.

\bibitem{fc3d}
H.~Fan, X.~Niu, Q.~Liu, and W.~Luk, ``F-c3d: Fpga-based 3-dimensional
  convolutional neural network,'' in {\em FPL'17}.

\bibitem{cambricon}
S.~Liu, Z.~Du, J.~Tao, D.~Han, T.~Luo, Y.~Xie, Y.~Chen, and T.~Chen,
  ``Cambricon: An instruction set architecture for neural networks,'' in {\em
  ACM SIGARCH Computer Architecture News}.

\bibitem{cnnopt}
C.~Zhang, P.~Li, G.~Sun, Y.~Guan, B.~Xiao, and J.~Cong, ``Optimizing fpga-based
  accelerator design for deep convolutional neural networks,'' in {\em
  FPGA'15}.

\bibitem{suda}
N.~Suda, V.~Chandra, G.~Dasika, A.~Mohanty, Y.~Ma, S.~Vrudhula, J.-s. Seo, and
  Y.~Cao, ``Throughput-optimized opencl-based fpga accelerator for large-scale
  convolutional neural networks,'' in {\em FPGA'16}.

\bibitem{looporder}
Y.~Ma, Y.~Cao, S.~Vrudhula, and J.-s. Seo, ``Optimizing loop operation and
  dataflow in fpga acceleration of deep convolutional neural networks,'' in
  {\em FPGA'17}.

\end{thebibliography}

\end{document}